\documentclass{article}

\usepackage{xcolor}
\usepackage{bm}

\usepackage[preprint]{corl_2026} 

\usepackage{graphicx,tikz}
\usepackage{amsmath,amssymb}
\usepackage{booktabs,wrapfig}
\usepackage{capt-of}
\usepackage{placeins}
\usepackage{needspace}
\usepackage{etoolbox}
\usetikzlibrary{arrows.meta}
\DeclareRobustCommand{\systemname}{UMI-Live}

\makeatletter
\patchcmd{\@maketitle}{\vskip 0.3in \@minus 0.1in}{\vskip 0.12in}{}{\PackageError{umi-live}{Could not adjust title spacing}{}}
\makeatother

\title{UMI-Live for In-The-Wild Bimanual Robotics Data Collection}

\author{%
  {\large\bfseries Govind Velamoor$^{1,}$\thanks{Equal first-author contribution.}\quad
  Adrian Lin$^{2,*}$\quad Sam Jung$^{3}$}\\[1pt]
  {\large\bfseries Spencer Mateega$^{3}$\quad
  Brandon Y.~Yang$^{4,}$\thanks{Equal advising.}\quad
  Amish Sethi$^{5,\dagger,}$\thanks{Corresponding author. \texttt{amishsethi@g.harvard.edu}.}}\\[5pt]
  {\normalfont\normalsize $^{1}$Yale University\quad $^{2}$Carnegie Mellon University\quad $^{3}$AfterQuery}\\
  {\normalfont\normalsize $^{4}$University of Pennsylvania\quad $^{5}$Harvard University}
}

\hypersetup{pdftitle={UMI-Live for In-The-Wild Bimanual Robotics Data Collection},
            pdfauthor={Govind Velamoor, Adrian Lin, Sam Jung, Spencer Mateega, Brandon Y. Yang, Amish Sethi}}

\begin{document}
\maketitle

\begin{abstract}
Universal Manipulation Interface (UMI) devices allow human experts to demonstrate tasks for robot learning without teleoperating physical robots. Using these demonstrations as training data still requires a consistent hardware setup, calibration, and data processing. We present UMI-Live, a full-stack open-source bimanual UMI system for in-the-wild data collection. We design our custom UMI-Live hardware to integrate camera and LiDAR--inertial sensing, onboard compute, and power, without external tracking equipment or calibration within the collection environment. We implement a simple user interface for recording data while an asynchronous pipeline performs parallel SLAM reconstruction and publishes training-ready datasets. We also release software that supports dual-YAM teleoperation and VLA policy training and inference. We validate UMI-Live data quality by fine-tuning $\pi_{0.5}$ and MolmoAct2, achieving success rates of 10/10 and 8/10, respectively on a bimanual pick-and-place task. Furthermore, we release 40 hours of UMI-Live data across three tasks, and complete hardware designs, software, and reproduction guides at our \href{https://umi-live.github.io/}{project website}.
\end{abstract}

\begin{figure}[!htbp]
\centering
\resizebox{\linewidth}{!}{\input{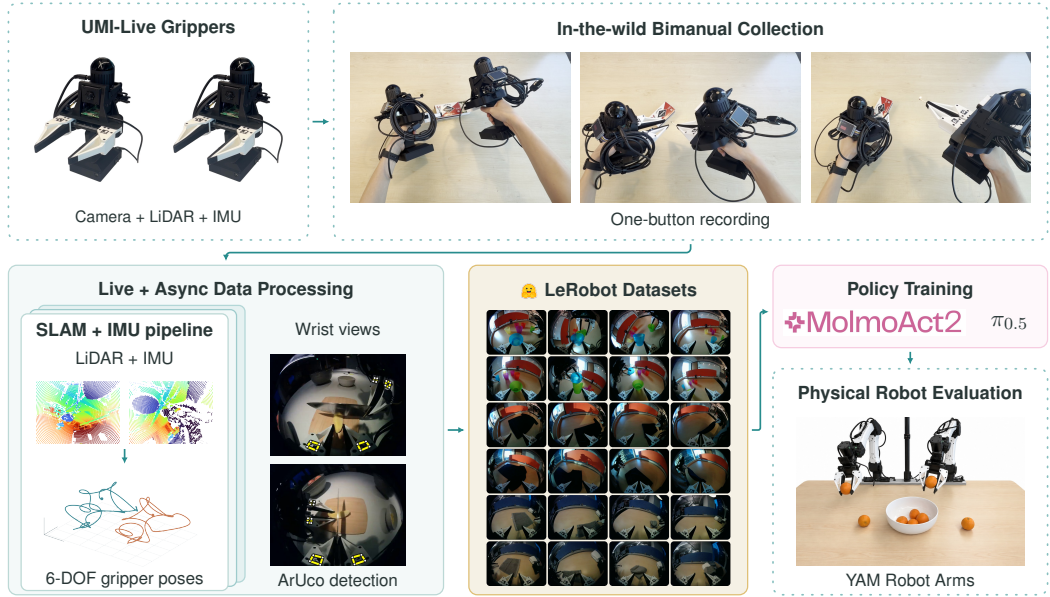}}
\caption{UMI-Live connects in-the-wild bimanual collection and automated processing to LeRobot datasets, policy training, and physical evaluation on dual i2RT YAM arms.}
\label{fig:overview}
\end{figure}
\FloatBarrier


\begin{figure}[!htbp]
\centering
\resizebox{\linewidth}{!}{\input{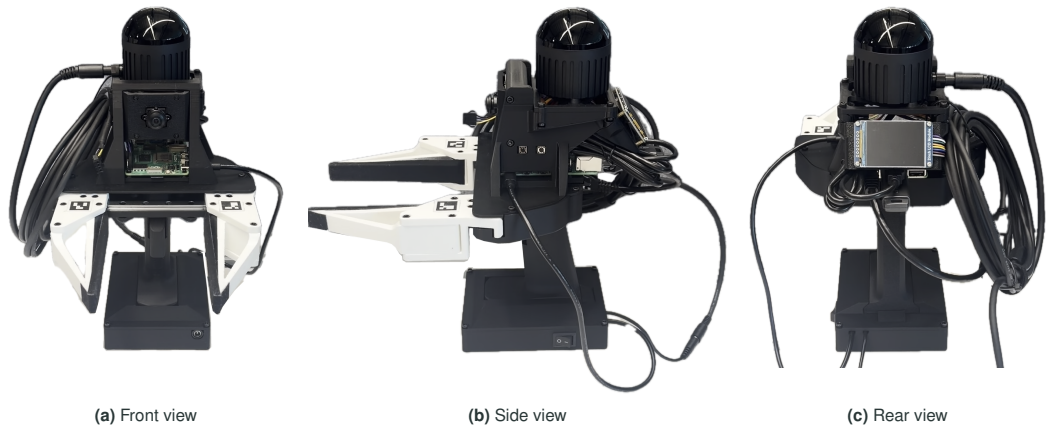}}
\caption{Assembled \systemname{} hardware. \textbf{(a)} Front view showing the wrist camera and ArUco tags. \textbf{(b)} Side view showing the operator-facing buttons and battery module. \textbf{(c)} Rear view showing the LCD and cable routing.}
\label{fig:hardware-views}
\end{figure}

\section{Introduction}

Much of the recent progress in robot manipulation has come from imitation learning, in which policies learn manipulation behaviors from human demonstrations, increasingly drawn from large and diverse datasets. Diffusion policies~\citep{chi2024diffusionpolicyvisuomotorpolicy} and vision-language-action (VLA) models such as RT-2, OpenVLA, $\pi_0$, $\pi_{0.5}$, and GR00T N1~\citep{brohan2023rt2visionlanguageactionmodelstransfer,kim2024openvlaopensourcevisionlanguageactionmodel,black2026pi0visionlanguageactionflowmodel,intelligence2025pi_,nvidia2025gr00tn1openfoundation} are all trained this way. More recently, world-action models (WAMs) jointly predict future visual observations and robot actions~\citep{zhu2025unifiedworldmodelscoupling,kim2026cosmospolicyfinetuningvideo,ye2026worldactionmodelszeroshot}, which allows them to build on large-scale video pretraining. Video alone, however, carries no robot action labels, so WAMs still learn to act from action-labeled demonstrations. These methods allow robots to perform tasks ranging from pick-and-place to dexterous manipulation, including towel and shirt folding~\citep{black2026pi0visionlanguageactionflowmodel}. Learning these skills in new environments still depends on expert demonstrations covering the relevant objects and interactions. Collecting high-quality demonstrations across diverse environments therefore continues to be a key challenge in robot learning.

Action-labeled demonstrations are commonly collected by teleoperating physical robots. Large collection efforts such as DROID and ABC-130K show the value of collecting demonstrations across operators, tasks, and environments~\citep{khazatsky2025droidlargescaleinthewildrobot,allshire2026scalablebehaviorcloningopen}. Because each collection site must host the robot along with its control, perception, and safety infrastructure, expanding to new environments requires transporting this equipment  or maintaining additional robot stations. The Universal Manipulation Interface (UMI) removes this requirement by letting experts demonstrate tasks with handheld grippers instead of robots~\citep{chi2024universal}. UMI showed that policies trained on wrist observations and reconstructed gripper trajectories from in-the-wild demonstrations can perform challenging tasks, including bimanual ones.

Subsequent UMI systems have explored different sensing, tracking, and gripper designs. However, reproducing these systems requires clear instructions for building and calibrating the hardware, together with software for processing the recordings. Some UMI configurations use stationary tracking rigs or backpack-mounted computers, which add equipment to set up or carry~\citep{ohkawa2026yubiyieldinguniversalbidigital,zeng2025activeumiroboticmanipulationactive}. Bulky gripper designs can also make dexterous manipulation of small objects difficult~\citep{ohkawa2026yubiyieldinguniversalbidigital}. Furthermore, each recording must be reconstructed and converted into a training-ready dataset. Manually managing these steps becomes tedious, and we believe that operators should be able to start the next demonstration while the software prepares completed recordings for training and physical robot evaluation.

We present UMI-Live, a complete hardware and software system for in-the-wild bimanual collection and robot learning. Each UMI-Live gripper integrates camera and LiDAR--inertial sensing, onboard compute and storage, and removable power. An operator can continue recording data with the press of a single button, while an asynchronous pipeline transfers completed recordings and performs parallel reconstruction, gripper position and jaw opening estimation, validation, and dataset preparation. We also provide software for dual-YAM teleoperation, policy training, and inference. We release approximately 40 hours of demonstrations across three tasks and evaluate policies fine-tuned on mixtures containing mostly UMI demonstrations. The handheld demonstrations are collected at a different location and under different lighting conditions from those used for robot evaluation, while the smaller teleoperation component provides demonstrations from the target robot and scene.

We make the following contributions.
\begin{enumerate}
\item \textbf{Reproducible bimanual hardware for in-the-wild collection.} We present self-contained handheld grippers with integrated sensing, compute, storage, power, and operator controls, together with hardware designs and construction and calibration guides for reproducing the system.
\item \textbf{Complete software integration from recording to robot deployment.} We provide software for coordinated recording, asynchronous transfer, parallel and batched processing, LeRobot dataset preparation, dual-YAM teleoperation, and policy training and inference. Operators can continue collecting while completed demonstrations are processed.
\item \textbf{A public bimanual dataset and physical robot validation.} We release approximately 40 hours of demonstrations across three manipulation tasks and evaluate two VLA policies through physical rollouts. Following mixed-data fine-tuning, $\pi_{0.5}$ succeeds in up to 10/10 trials and MolmoAct2 succeeds in 8/10 trials compared with 0/10 for its released checkpoint.
\end{enumerate}

\section{Related Work}
\label{sec:related-work}

\paragraph{Robot demonstration collection.}
Teleoperation records observations and actions directly on the target embodiment. BridgeData V2, DROID, RoboMIND, ALOHA Unleashed, and ABC-130K illustrate the breadth of manipulation data that can be collected through physical robot operation~\citep{walke2024bridgedatav2datasetrobot,khazatsky2025droidlargescaleinthewildrobot,Wu_2025,zhao2024alohaunleashedsimplerecipe,allshire2026scalablebehaviorcloningopen}. These efforts cover different tasks, environments, and embodiments, but require a robot for demonstration capture. \systemname{} supports handheld collection independently of robot operation and provides a separate teleoperation configuration for obtaining complementary demonstrations on the target embodiment.

\paragraph{Handheld UMI systems.}
The original UMI combines a handheld gripper, wrist video, gripper-state sensing, and visual--inertial trajectory reconstruction to support robot learning from portable demonstrations~\citep{chi2024universal}. FastUMI develops the hardware, tracking, and collection tools for broader deployment~\citep{zhaxizhuoma2025fastumiscalablehardwareindependentuniversal}. UMI-3D introduces LiDAR-centered reconstruction and an integrated acquisition, calibration, training, and deployment pipeline, with a reported single-arm configuration~\citep{wang2026umi3dextendinguniversalmanipulation}. \systemname{} uses two independently instrumented LiDAR--inertial devices as one coordinated bimanual collection system.

Other systems make different choices about sensing and interaction. YUBI emphasizes finger-aligned operation and bimanual dexterity, with stationary and portable collection configurations~\citep{ohkawa2026yubiyieldinguniversalbidigital}. ActiveUMI combines a VR headset with a backpack-mounted computer to record active egocentric perception~\citep{zeng2025activeumiroboticmanipulationactive}. HiFi-UMI uses head-mounted stereo--inertial sensing, hardware synchronization, and direct relative-pose estimation to improve bimanual trajectory fidelity~\citep{ai2026hifiumilearningdeployablemanipulation}. \systemname{} places sensing, compute, storage, power, and recording controls on the handheld devices and connects their outputs to an automated processing system and a dual-YAM learning workflow.

\paragraph{Collection and robot-learning software.}
LeRobot provides shared dataset representations and tools for policy training, evaluation, and inference~\citep{cadene2026lerobotopensourcelibraryendtoend}. RoboPocket connects portable collection to remote policy feedback and asynchronous fine-tuning~\citep{fang2026robopocketimproverobotpolicies}, and ABC provides open infrastructure spanning collection, training, and evaluation~\citep{allshire2026scalablebehaviorcloningopen}. \systemname{} combines bimanual LiDAR--inertial recording, asynchronous reconstruction and LeRobot dataset publication, and dual-YAM teleoperation, policy training, and inference in a reproducible setup.


\begin{figure}[!t]
\centering
\resizebox{\linewidth}{!}{\input{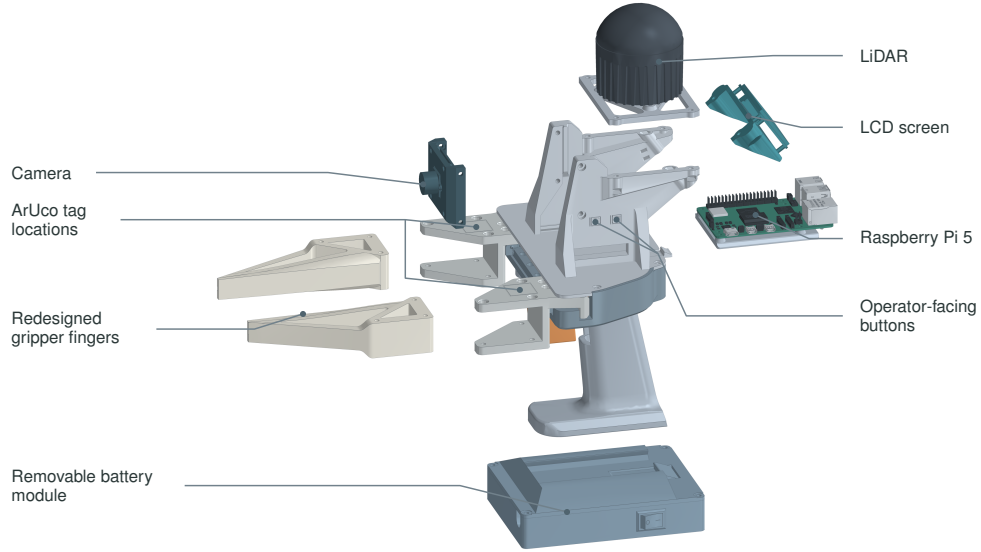}}
\caption{Exploded CAD view of one \systemname{} unit with its main components labeled.}
\label{fig:cad-modules}
\end{figure}

\section{System Design}

Figure~\ref{fig:overview} shows how \systemname{} connects in-the-wild bimanual collection to automated processing, LeRobot datasets, and physical robot learning. We first describe the handheld hardware, including the gripper mechanism, sensing, and power design (Section~\ref{sec:hardware}), followed by recording, trajectory reconstruction, dataset export, and parallel processing (Section~\ref{sec:pipeline}). We then evaluate policies trained on the collected demonstrations (Section~\ref{sec:experiments}).
\subsection{Hardware Design}
\label{sec:hardware}
\paragraph{Design objectives.}
\systemname{} hardware consists of two handheld grippers with onboard sensing, compute, and storage for bimanual data collection. Our design focuses on modularity, usability, and user-friendliness. Figure~\ref{fig:hardware-views} shows the assembled unit from the front, side, and rear, and Figure~\ref{fig:cad-modules} shows its main components. Appendix~\ref{app:hardware} provides additional hardware details.


\paragraph{Gripper mechanism.}
We improve the original UMI gripper mechanism~\citep{chi2024universal} by using bearing-supported trigger and gear pivots for smoother actuation and reduced backlash. Squeezing the trigger rotates two sector gears that engage racks on the finger mounts (Figure~\ref{fig:hardware-details}a,b). The mounts travel along a linear guide to produce opposing parallel-jaw motion. To achieve smoother gripper motion, we apply adhesive tape as shown in Figure~\ref{fig:hardware-details}. We also place ArUco tags~\citep{garridojurado2014aruco} on the moving finger mounts to measure the jaw opening.

\Needspace{16\baselineskip}
\begin{wrapfigure}[12]{R}{0.48\textwidth}
\centering
\resizebox{\linewidth}{!}{\input{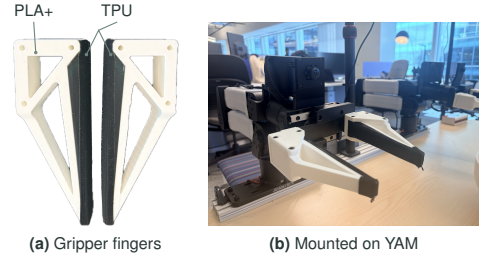}}
\caption{Redesigned gripper fingers and their installation on a YAM arm.}
\label{fig:gripper-fingers}
\end{wrapfigure}

\paragraph{Gripper finger design.}
We redesign the UMI gripper finger to adapt to more tasks that involve dexterous manipulation with smaller objects. We print the gripper fingers with a rigid PLA+ base and a thin TPU contact layer (Figure~\ref{fig:gripper-fingers}). The rigid base resists bending under grasping loads and it supports force application and precision contact with small objects such as screws. The TPU layer provides compliance at the contact surface, which helps the fingers conform to different object shapes. We demonstrate the capabilities of the redesigned grippers through human-operated tasks involving small-object grasping and insertion, match lighting, and shirt folding in Appendix~\ref{app:manipulation-demonstrations}.

\paragraph{Sensing and electronics.}
Each gripper carries a RoboSense Airy LiDAR with an integrated IMU\footnote{\url{https://store.robosense.ai/}}, a global-shutter wide-angle USB camera\footnote{SVPRO AR0234 global-shutter fisheye USB camera: \url{https://www.amazon.com/dp/B0GYRSG984}}, and a Raspberry Pi 5 (Figures~\ref{fig:cad-modules} and~\ref{fig:hardware-details}d). The camera records wrist views at $800\times600$ pixels and 30 fps. We choose to use an ultrawide camera to capture the workspace and relevant global context as we do not use an overhead top camera.

\paragraph{Power.}
\begin{wraptable}{R}{0.43\textwidth}
\centering
\begin{tabular}{lr}
\toprule
Specification & Value \\
\midrule
Mass per unit with battery & 1.42\,kg \\
Gripper width & 80\,mm \\
Runtime per charge & 1.5\,h \\
Hardware cost per pair & \$1{,}930 \\
\bottomrule
\end{tabular}
\caption{Hardware specifications of the evaluated revision.}
\label{tab:hardware-specifications}
\end{wraptable}
We use a removable battery module that clips to the base of each handle (Figure~\ref{fig:hardware-details}c). The module houses the battery, power switch, and DC-DC converter, which supplies 5\,V USB-C power to the Raspberry Pi and 12\,V to the RoboSense Airy LiDAR.

\paragraph{In-the-wild operation.}
Operators start and stop recording on both grippers with a single button, and the LCD shows recording and transfer status. Operators can collect in the wild without a robot, tripod-mounted cameras, external tracking equipment, or environment specific calibration. Table~\ref{tab:hardware-specifications} lists the main specifications.

\subsection{Operator Interface and Recording}
\label{sec:recording}

\begin{wrapfigure}{l}{0.27\textwidth}
\vspace{-\intextsep}
\centering
\includegraphics[width=\linewidth]{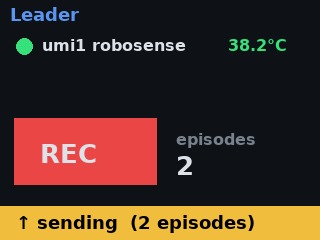}
\caption{LCD interface with example values.}
\label{fig:lcd-interface}
\end{wrapfigure}
The operator starts and stops recording on both grippers with a single button. The LCD shows recording status and transfer progress (Figure~\ref{fig:lcd-interface}).

For episode $e$, each gripper $i\in\{L,R\}$ stores timestamped wrist video and raw LiDAR and IMU packets locally on its Raspberry Pi as $\mathcal R_i^{(e)}$. Together, the two recordings form $\mathcal R^{(e)}=(\mathcal R_L^{(e)},\mathcal R_R^{(e)})$. Completed recordings are transferred to the server and processed asynchronously, which allows the operator to continue collecting demonstrations.
\WFclear

\begin{figure}[!t]
\centering
\resizebox{\linewidth}{!}{\input{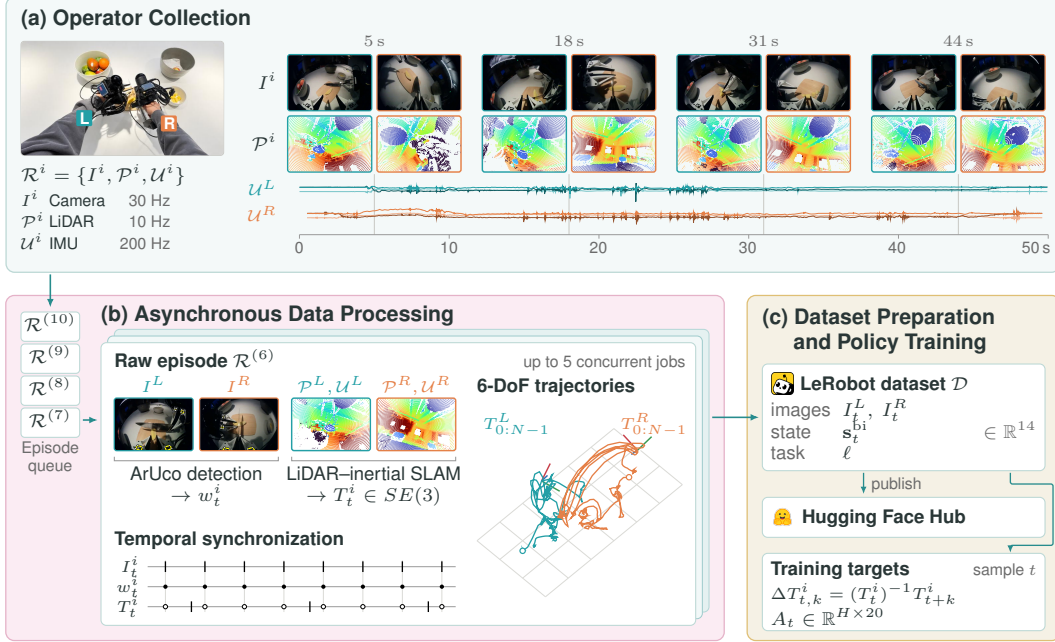}}
\caption{Data collection and processing in \systemname{}. \textbf{(a)} Each gripper \(i\in\{L,R\}\) records wrist images \(I^i\), LiDAR data \(\mathcal P^i\), and IMU data \(\mathcal U^i\). \textbf{(b)} Raw episodes \(\mathcal R^{(e)}\) are processed in parallel to estimate jaw widths \(w_t^i\), reconstruct poses \(T_t^i\), and align the streams. \textbf{(c)} We assemble demonstration trajectories \(\mathcal T^{(e)}\) with wrist images, 14-dimensional bimanual states, and task instructions into dataset \(\mathcal D\). Training samples use observations \(o_t=(I_t^L,I_t^R,g_t^L,g_t^R,\ell)\) and query-relative action chunks \(A_t\in\mathbb R^{H\times20}\), as defined in Appendices~\ref{app:dataset-representation} and~\ref{app:policy-training}.}
\label{fig:software-pipeline}
\end{figure}

\subsection{Automated Data Processing}
\label{sec:pipeline}

Once the completed recordings reach the server, they enter the processing pipeline shown in Figure~\ref{fig:software-pipeline}. The server validates synchronization, estimates TCP trajectories and gripper openings, and combines these measurements with wrist images into episodes ready for training.

\paragraph{Synchronization and time correction.}
The server first checks the synchronization evidence saved during recording and excludes episodes with unacceptable sensor synchronization or an estimated cross-device offset above 5\,ms. During recording, Chrony synchronizes the Raspberry Pi clocks, and gPTP synchronizes each Airy sensor with its connected Pi. Camera exposure timestamps are mapped from the Pi's monotonic clock to host time. The server then derives an affine mapping between embedded LiDAR timestamps and Pi-side packet-arrival timestamps. The corrected timestamps are used when reconstructed poses are aligned with camera observations.

\paragraph{Pose and gripper estimation.}
The server decodes each device's LiDAR and IMU packets and runs GLIM~\citep{Koide_2024} independently for the two grippers. The tool center point (TCP) is defined at the center of the gripper tips. A fixed sensor-to-TCP offset converts the reconstructed LiDAR body motion into a TCP trajectory. Appendix~\ref{app:yam-kinematics} specifies the corresponding TCP frame and flange-to-TCP transform used on the YAM arms.

Jaw opening is estimated from the two ArUco tags attached to each gripper's moving finger mounts. The server unprojects tag centers through the calibrated fisheye model and combines their angular separation with the fixed camera-to-tag geometry to estimate physical width. Missing detections are interpolated between valid measurements. Calibrated closed and open endpoints map these widths to normalized gripper values.

\paragraph{Temporal alignment and dataset assembly.}
The leader camera defines the episode timeline. We interpolate reconstructed positions linearly and rotations through quaternion spherical interpolation at these frame times. The follower video contributes the latest available frame at or before each corresponding time, and gripper openings are interpolated onto the same timeline.

Each hand's TCP pose is expressed relative to its own fixed episode-start pose. At timestep $t$, its state is
\[
\mathbf s_t^i=[\mathbf p_t^i;\boldsymbol\phi_t^i;g_t^i]\in\mathbb R^7,
\]
where $\mathbf p_t^i$ is position in meters, $\boldsymbol\phi_t^i$ is a rotation vector in radians, and $g_t^i\in[0,1]$ is normalized gripper opening. The bimanual state is
\[
\mathbf s_t^{\mathrm{bi}}=[\mathbf s_t^L;\mathbf s_t^R]\in\mathbb R^{14}.
\]
An episode with $N$ frames and task instruction $\ell$ becomes
\[
\mathcal T=
\left[\left(I_t^L,I_t^R,\mathbf s_t^{\mathrm{bi}},\ell\right)\right]_{t=0}^{N-1}.
\]
We collect these trajectories into $\mathcal D=\{\mathcal T^{(e)}\}_{e=1}^{M}$, where $M$ is the number of episodes.

The software checks stream integrity, synchronization evidence, required pose and gripper measurements, frame counts, and dataset consistency before accepting an episode. Finalized episodes are stored in LeRobot format. For cloud publication, the software synchronizes the dataset and raw-episode archive with Hugging Face. The software retains the local dataset if publication fails.

\paragraph{Parallel execution.}
Processing follows a dependency graph across hands and episodes. LiDAR decoding, SLAM, and gripper estimation execute concurrently when their inputs are available. Dataset assembly waits for the required outputs and uses a serialized writer. The software tracks completed stages so interrupted processing can resume from available outputs.

\subsection{Policy Training and Deployment}
\label{sec:policy-training}

We fine-tune $\pi_{0.5}$ and MolmoAct2 on UMI and robot teleoperation demonstrations using the same observation and action representation. The policies take two wrist images, gripper openings, and a task instruction as input. They predict action chunks containing query-relative TCP poses and absolute gripper openings. Appendix~\ref{app:policy-training} describes the representation, training objective, and settings.

During deployment, a remote policy server returns action chunks to the local dual-YAM client, which converts them into joint and gripper commands and coordinates the two arms. Appendix~\ref{app:policy-inference} describes inference and the kinematic model.

\section{Experimental Results}
\label{sec:result}
\label{sec:experiments}

We evaluate whether demonstrations collected with \systemname{} support bimanual policy learning and physical robot execution. Our evaluation uses an orange-to-bowl task on dual YAM arms. We fine-tune $\pi_{0.5}$~\citep{intelligence2025pi_} and MolmoAct2~\citep{fang2026molmoact2} using mixtures containing mostly UMI demonstrations and compare the fine-tuned MolmoAct2 policy with its released checkpoint.

\begin{figure}[!htbp]
\centering
\edef\experimentpanelheight{\the\dimexpr0.34\linewidth\relax}
\begin{minipage}[t]{0.44\linewidth}
\centering
\begin{minipage}[c][\experimentpanelheight][c]{\linewidth}
\includegraphics[width=0.49\linewidth]{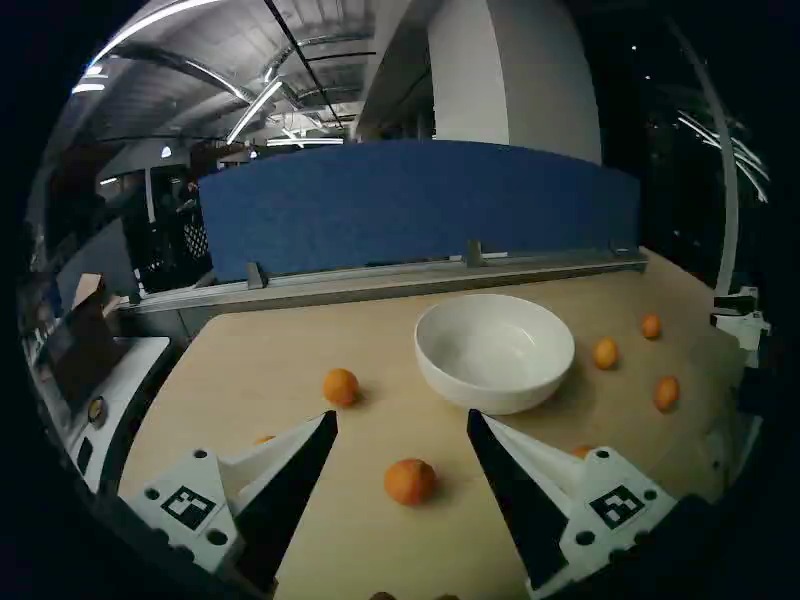}\hfill
\includegraphics[width=0.49\linewidth]{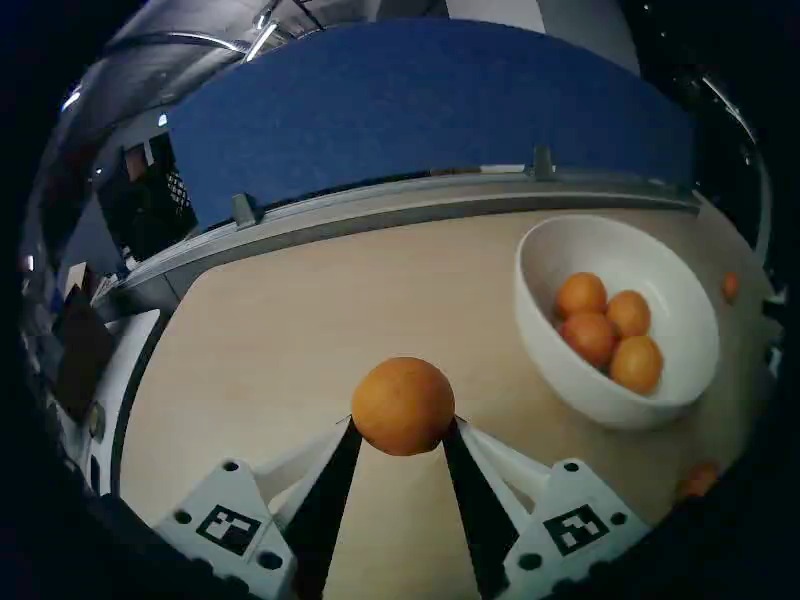}\\[2pt]
\includegraphics[width=0.49\linewidth]{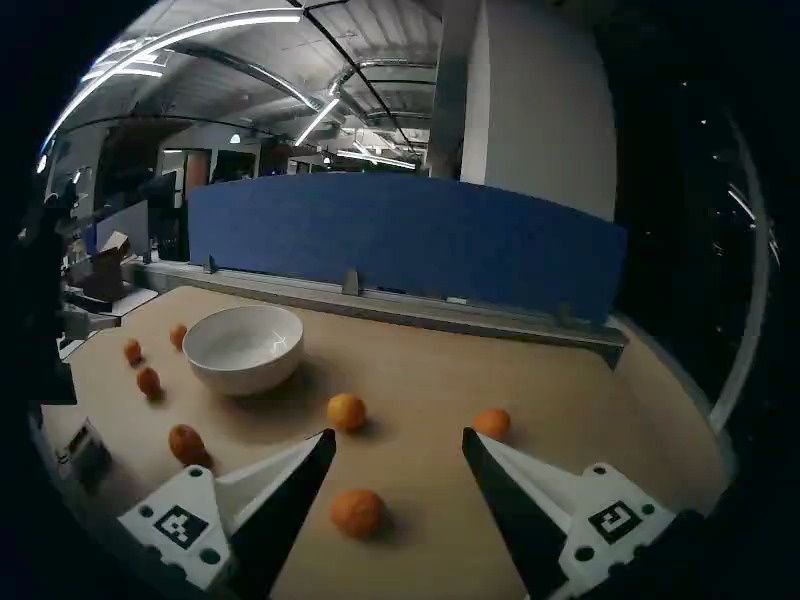}\hfill
\includegraphics[width=0.49\linewidth]{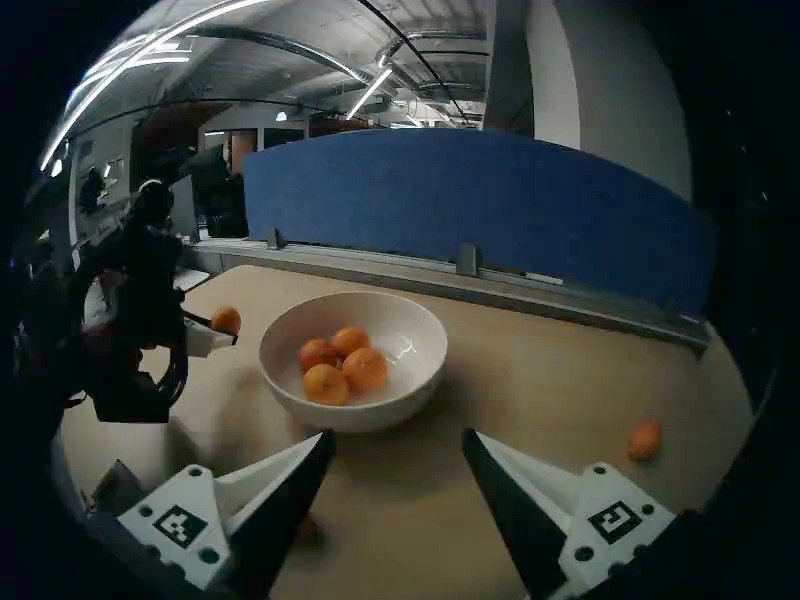}
\end{minipage}\\[3pt]
{\small\textbf{(a)} Handheld UMI collection}
\end{minipage}\hspace{0.06\linewidth}%
\begin{minipage}[t]{0.42\linewidth}
\centering
\begin{minipage}[c][\experimentpanelheight][c]{\linewidth}
\includegraphics[width=\linewidth]{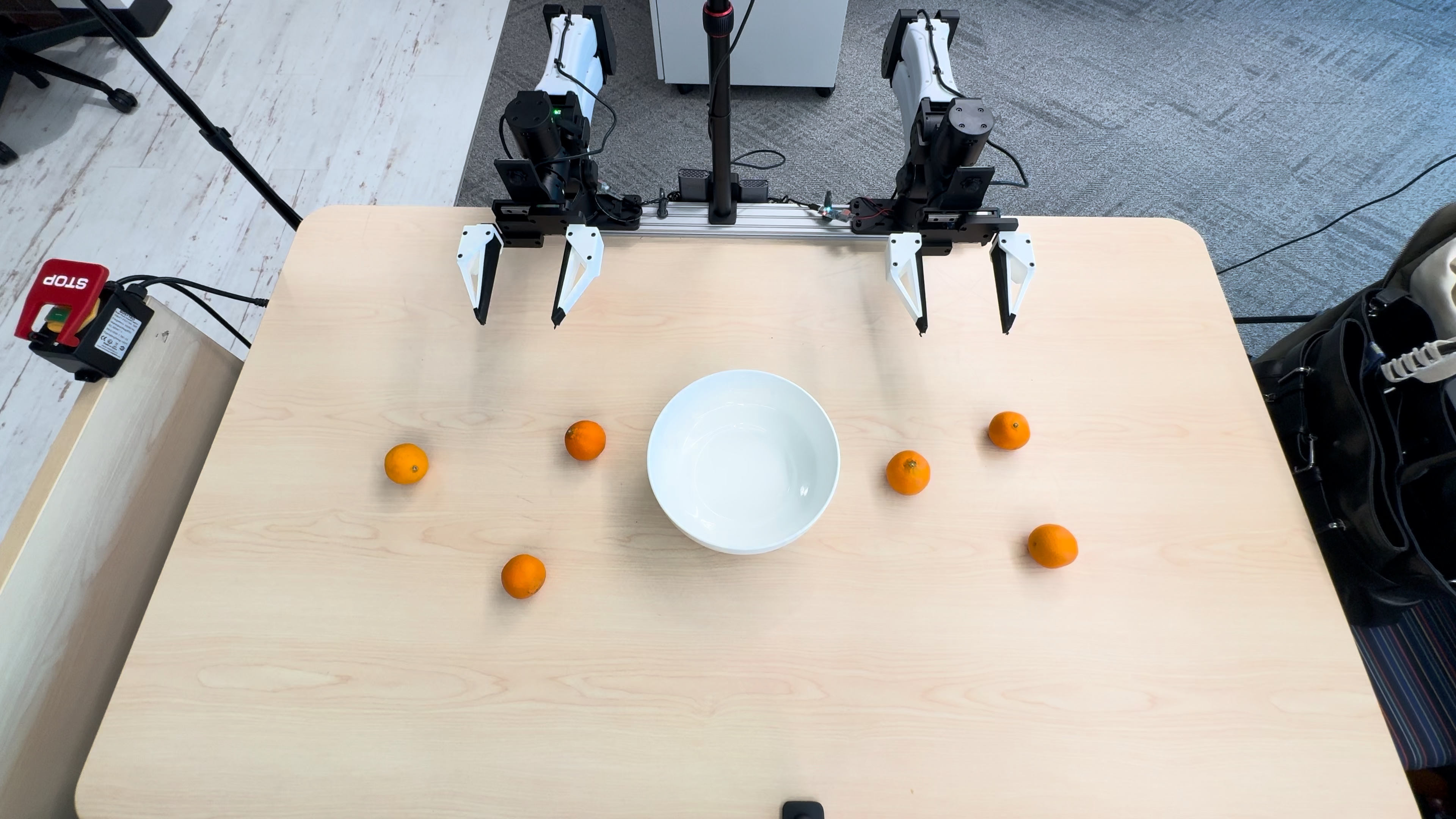}
\end{minipage}\\[3pt]
{\small\textbf{(b)} Robot evaluation setup}
\end{minipage}\par
\vspace{6pt}
\includegraphics[width=0.245\linewidth]{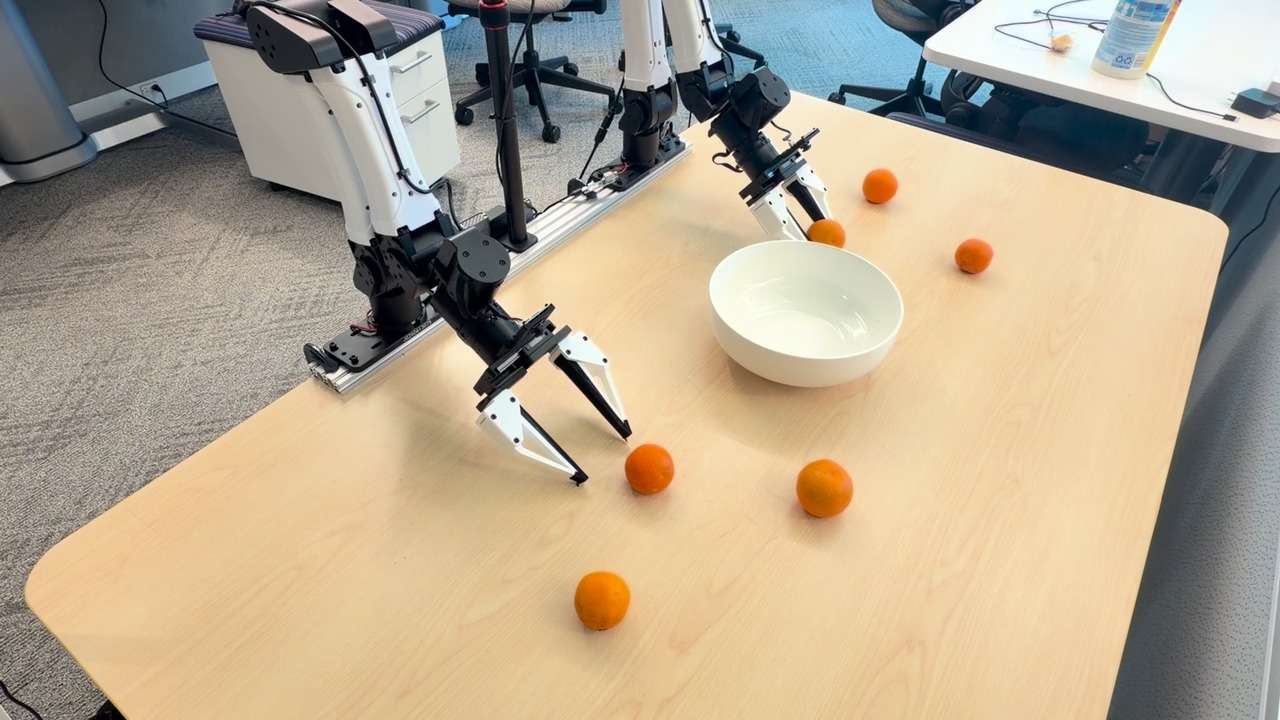}\hfill
\includegraphics[width=0.245\linewidth]{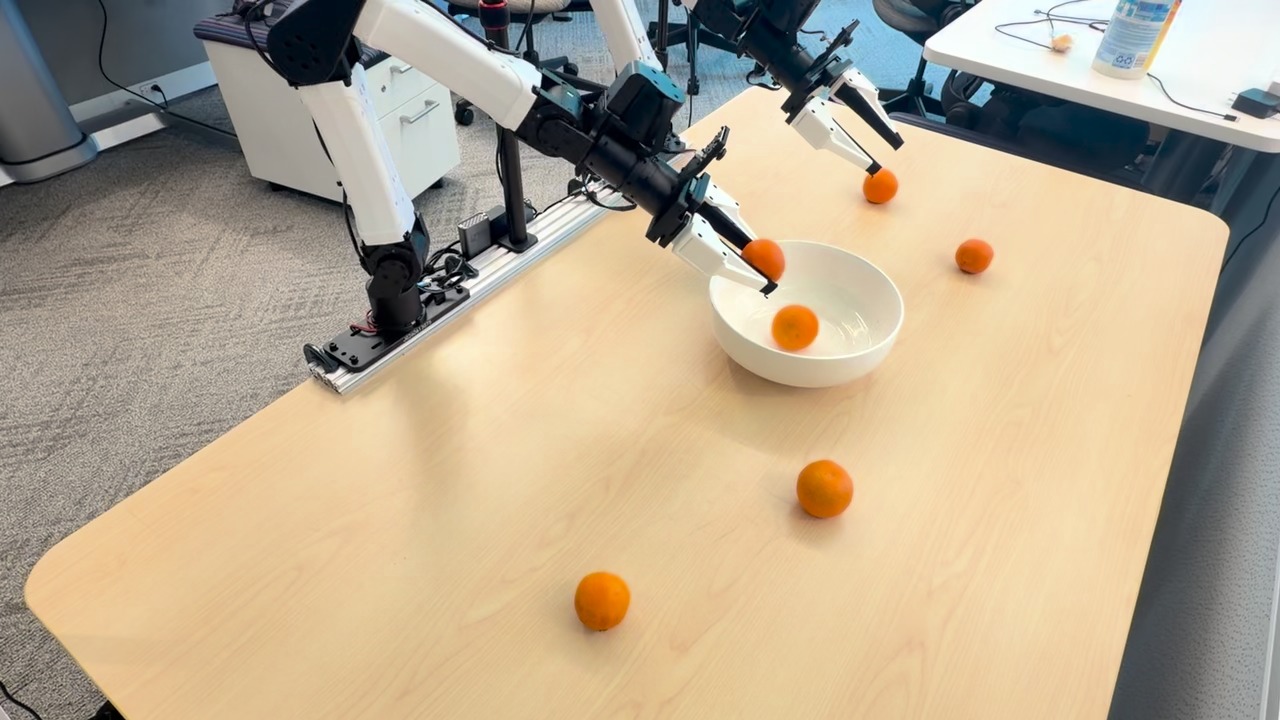}\hfill
\includegraphics[width=0.245\linewidth]{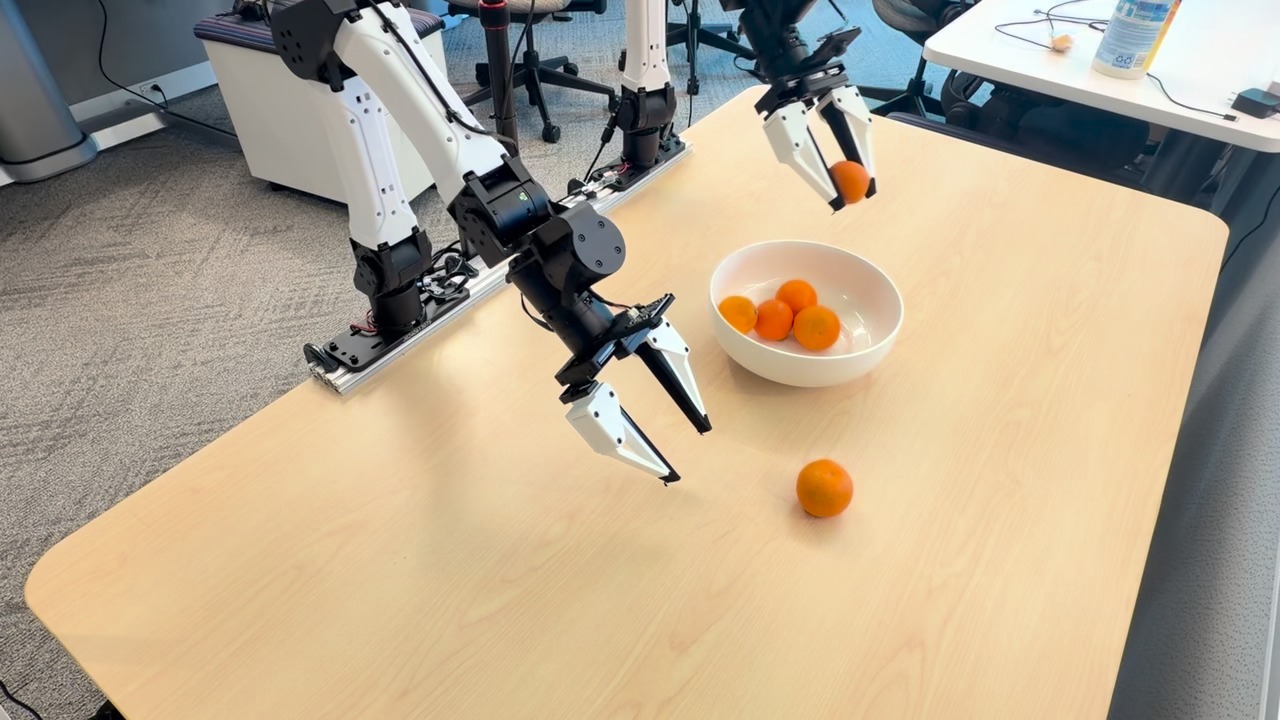}\hfill
\includegraphics[width=0.245\linewidth]{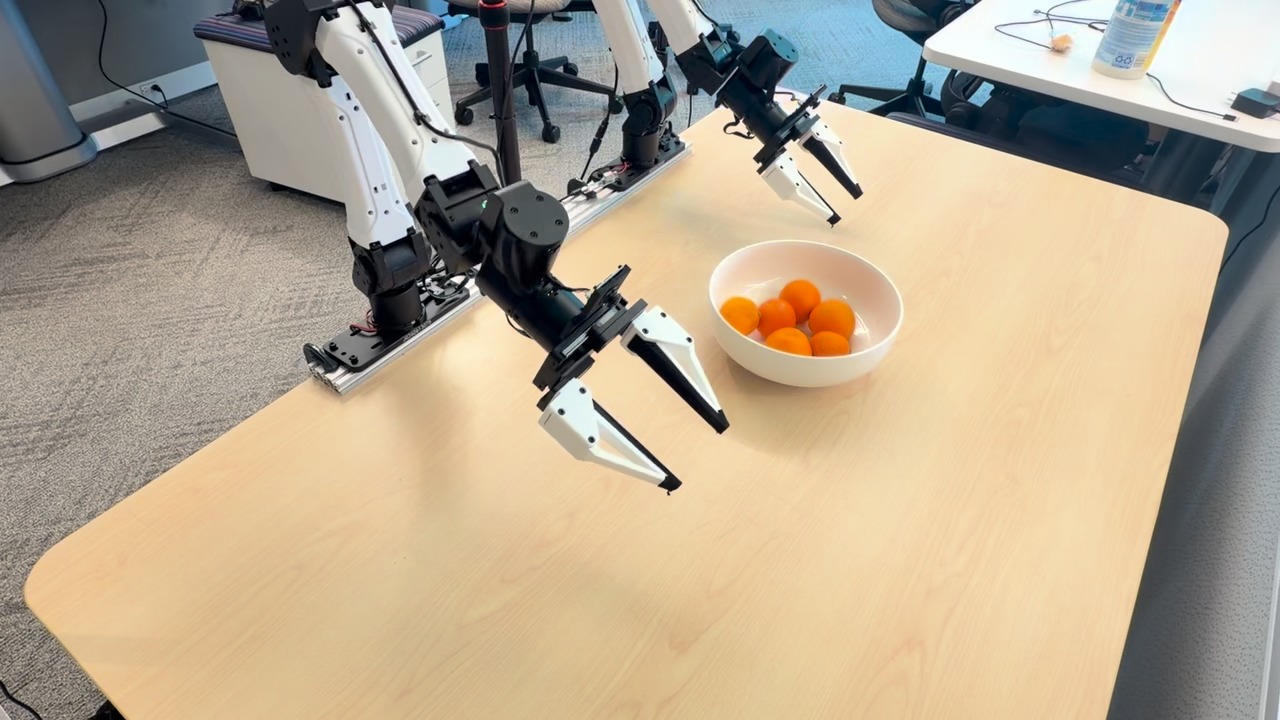}\\[3pt]
{\small\textbf{(c)} Autonomous $\pi_{0.5}$ rollout}
\caption{Handheld collection and physical robot evaluation. \textbf{(a)} Wrist-camera views at two moments in a UMI demonstration collected at a different location and under different lighting from the robot evaluation, brightened for display. \textbf{(b)} Dual i2RT YAM arms with UMI-Live gripper fingers, six oranges, and a bowl. \textbf{(c)} Selected frames from one autonomous $\pi_{0.5}$ rollout that ends with all six oranges in the bowl. Frames are ordered from left to right. Appendix~\ref{app:autonomous-rollout}, Figure~\ref{fig:autonomous-rollout}, shows the full sequence.}
\label{fig:experiment-setup}
\label{fig:experiment-rollout}
\end{figure}

\subsection{Experimental Setup}
In our task, the robot must pick up six oranges from a table and place them in a bowl (Figure~\ref{fig:experiment-setup}b). Each trial starts from a fixed robot configuration, with randomized orange and bowl positions. A trial succeeds if all six oranges are placed in the bowl and fails if any remain outside the bowl when the trial ends. We allow grasp retries and recovery of dropped oranges. We conduct ten trials for each evaluated condition.

The UMI demonstrations were collected at a different location and under different lighting from the robot evaluation (Figure~\ref{fig:experiment-setup}a). Robot teleoperation demonstrations were collected in the evaluation scene using the setup described in Appendix~\ref{app:teleop-setup}. The experiments therefore evaluate learning from handheld demonstrations supplemented with robot data from the target environment.

During policy execution, the robot supplies two ultrawide wrist-camera images, the current gripper openings, and a task instruction. TCP poses and joint angles are omitted from the policy inputs. The policies predict bimanual action chunks containing relative TCP poses and absolute gripper openings. The execution controller maps these predictions to robot joint and gripper commands, as described in Appendix~\ref{app:policy-inference}.

\subsection{Training Conditions}
The training sources contain 296 UMI demonstrations and 75 robot teleoperation demonstrations of the orange-to-bowl task. These counts describe the training pool for this task, while the full approximately 40-hour dataset release spans three manipulation tasks.

We evaluate $\pi_{0.5}$ with UMI/teleoperation data splits of 95\%/5\% and 90\%/10\%, and MolmoAct2 with a split of 75\%/25\%. Each split specifies the UMI and robot teleoperation shares of the fine-tuning data, which sum to 100\%. Both sources are converted to the common observation and query-relative action representation described in Appendix~\ref{app:mixed-data-training}.

We initialize each $\pi_{0.5}$ run independently from its released base checkpoint and MolmoAct2 from the released MolmoAct2-BimanualYAM checkpoint. We update all $\pi_{0.5}$ parameters. For MolmoAct2, we train LoRA~\citep{hu2022lora} adapters in the vision-language backbone and the full action expert and keep the remaining backbone parameters frozen. We evaluate the final checkpoint from each run and use one training seed per condition. Appendix~\ref{app:training-setup} and Tables~\ref{tab:training-pi05} and~\ref{tab:training-molmoact2} provide the remaining hyperparameters.

As a baseline, we evaluate the released MolmoAct2-BimanualYAM checkpoint without additional training on our demonstrations.

\subsection{Results}
Table~\ref{tab:policy} reports the results. $\pi_{0.5}$ succeeds in nine of ten trials with the 95/5 mixture and all ten trials with the 90/10 mixture. MolmoAct2 succeeds in eight of ten trials after fine-tuning with the 75/25 mixture, compared with zero successful trials for its released checkpoint. Figure~\ref{fig:experiment-rollout}c shows selected frames from one autonomous $\pi_{0.5}$ execution. The full sequence appears in Appendix~\ref{app:autonomous-rollout}, Figure~\ref{fig:autonomous-rollout}.

\noindent\begin{minipage}{\linewidth}
\centering
\begin{tabular*}{\linewidth}{@{\extracolsep{\fill}}llcc@{}}
\toprule
Policy & Training condition & UMI / teleoperation & Successful trials \\
\midrule
MolmoAct2 & Released checkpoint & --- & 0/10 \\
$\pi_{0.5}$ & Fine-tuned & 95\% / 5\% & 9/10 \\
$\pi_{0.5}$ & Fine-tuned & 90\% / 10\% & 10/10 \\
MolmoAct2 & Fine-tuned & 75\% / 25\% & 8/10 \\
\bottomrule
\end{tabular*}
\captionof{table}{Physical orange-to-bowl evaluation on dual i2RT YAM arms. Every fine-tuned condition starts from its released checkpoint and receives 12,000 updates with a global batch size of 32. Success requires placing all six oranges in the bowl.}
\label{tab:policy}
\end{minipage}
\par\medskip

These results demonstrate successful physical execution after fine-tuning on mixtures containing mostly \systemname{} demonstrations. The handheld demonstrations were collected outside the evaluation environment, while the smaller teleoperation component supplies observations and motions from the target robot and scene. The evaluation covers one task and one training seed per condition.

\FloatBarrier


\section{Released Dataset}
\label{sec:released-dataset}

We release 3,255 bimanual episodes comprising approximately 39.7 hours at 30\,Hz. The release covers orange placement, water pouring, and towel folding, with synchronized wrist videos, bimanual pose and gripper states, and task instructions. The complete dataset is available on \href{https://huggingface.co/datasets/brandonyang/umi-live-dataset}{Hugging Face}.

Each hand uses a fixed episode-start reference, and we provide source provenance in the release. The standardized action field stores the same-frame state. Training loaders select future frames and derive the query-relative targets described in Appendix~\ref{app:policy-training}.


\section{Discussion and Limitations}

\label{sec:limitations}

\systemname{} spans handheld collection, automatic data preparation, and robot execution, and enables operators to continue collecting while completed recordings are simultaneously processed. However, \systemname{} devices still require one-time camera and gripper calibration. Additionally, the quality of LiDAR--inertial reconstruction depends on scene geometry and motion. Integrating sensing, compute, storage, and power increases mass to 1.42\,kg per unit. Our robot evaluation covers one task, ten trials per condition, and one training seed per condition, using mixtures of UMI and teleoperation data. Without UMI-only and teleoperation-only comparisons, we do not isolate the contribution of either source. Further evaluation should measure trajectory accuracy, processing latency and throughput, operator efficiency during extended collection, and policy performance across more tasks, training seeds, and data mixtures.

\section{Conclusion}

\label{sec:conclusion}

We present \systemname{}, a bimanual collection and robot-learning system that combines self-contained handheld hardware with automated data preparation and dual-YAM software integration. Operators control recording through a single interface while completed episodes undergo parallel processing. The resulting demonstrations connect to policy training and physical deployment through a common task-space representation.

We release approximately 40 hours of demonstrations across three tasks. Physical experiments show successful orange-to-bowl execution after fine-tuning $\pi_{0.5}$ and MolmoAct2 on mixtures containing mostly \systemname{} demonstrations. Altogether, our hardware, software, dataset, and robot evaluation compose a reproducible system spanning from handheld demonstration to physical robot learning.


\clearpage

\section*{Use of Generative AI}
The authors used LLMs to help with coding implementation and as a writing and editing aid for this manuscript. The authors take full responsibility for its content.

\acknowledgments{This material is based upon work supported by the U.S. National Science Foundation Graduate Research Fellowship Program under Grant No.~DGE-2630152. Any opinions, findings, and conclusions or recommendations expressed in this material are those of the author(s) and do not necessarily reflect the views of the National Science Foundation. This work received no other third-party funding. The authors declare no competing interests.

}


\bibliography{example}  

\clearpage
\appendix
\section{Hardware Details}
\label{app:hardware}
\begin{figure}[!htbp]
  \centering
  \includegraphics[width=\linewidth]{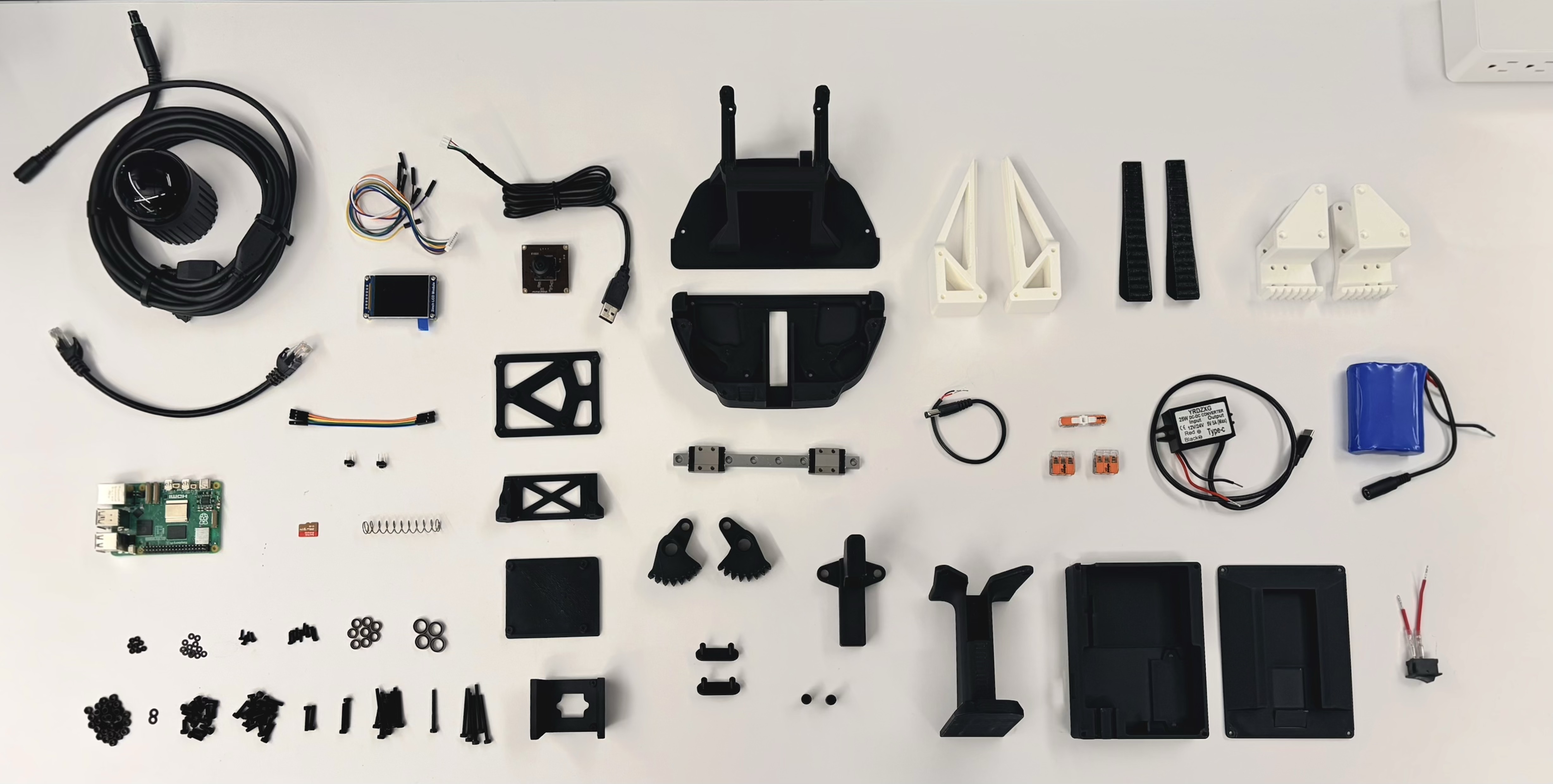}
  \caption{Components used to assemble a \systemname{} gripper, including the LiDAR, camera, Raspberry Pi, LCD, printed parts, linear guide, gears, fasteners, and battery and power electronics.}
  \label{fig:hardware-components}
\end{figure}

\begin{figure}[!htbp]
  \centering
  \includegraphics[width=\linewidth]{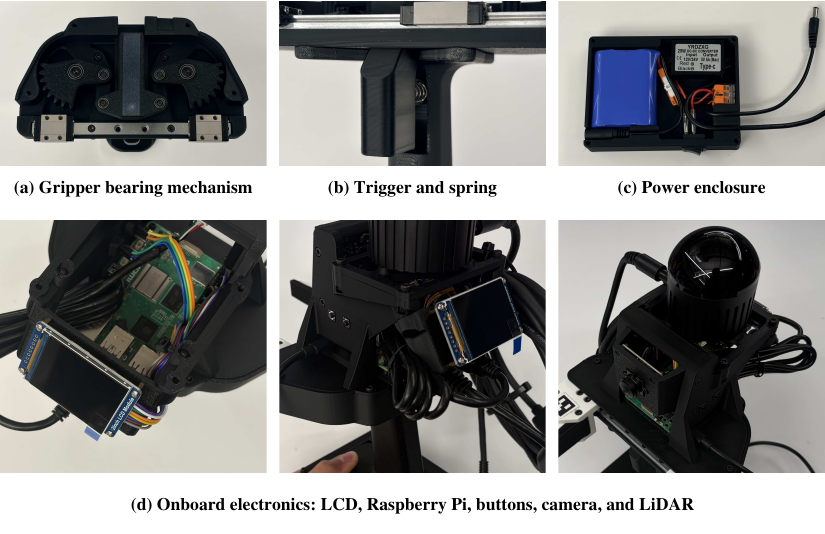}
  \caption{Mechanical and electronic components of \systemname{}. \textbf{(a)} Gripper mechanism with sector gears, bearing-supported pivots, and a linear guide. \textbf{(b)} Trigger and return spring within the handle. \textbf{(c)} Removable power module with the battery, DC-DC converter, power switch, and wiring. \textbf{(d)} Onboard electronics, showing the Raspberry Pi and LCD before LiDAR installation (left), operator-facing buttons and the rear-mounted LCD (center), and the front camera beneath the LiDAR (right).}
  \label{fig:hardware-details}
\end{figure}

Figure~\ref{fig:hardware-views} shows the assembled device. Figure~\ref{fig:hardware-components} shows the components used to assemble one unit. Figure~\ref{fig:hardware-details} shows the gripper mechanism, power module, and onboard electronics.

\FloatBarrier
\section{UMI-Live Manipulation Demonstrations}
\label{app:manipulation-demonstrations}
We demonstrate the manipulation capabilities of \systemname{} through three human-operated tasks involving small-object grasping and insertion, match lighting, and shirt folding.

\subsection{Small-Object Grasping and Insertion}
\label{app:precision-grasping}
\begin{figure}[!htbp]
  \centering
  \resizebox{\linewidth}{!}{\input{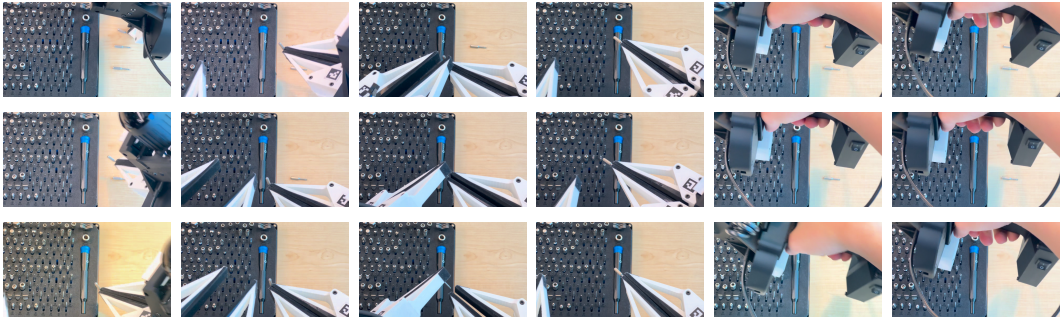}}
  \caption{Small-object grasping and insertion with \systemname{}. Each row begins at tool pickup and follows the handoff between the grippers, insertion into the tool case, and release. Six uncropped frames per row are selected at key moments and shown in chronological order from left to right.}
  \label{fig:precision-grasping}
\end{figure}

Figure~\ref{fig:precision-grasping} shows three manually operated sequences using \systemname{} to grasp small screwdriver bits and insert them into the tool case.

\FloatBarrier
\subsection{Match Lighting}
\label{app:match-lighting}
\begin{figure}[!htbp]
  \centering
  \resizebox{\linewidth}{!}{\input{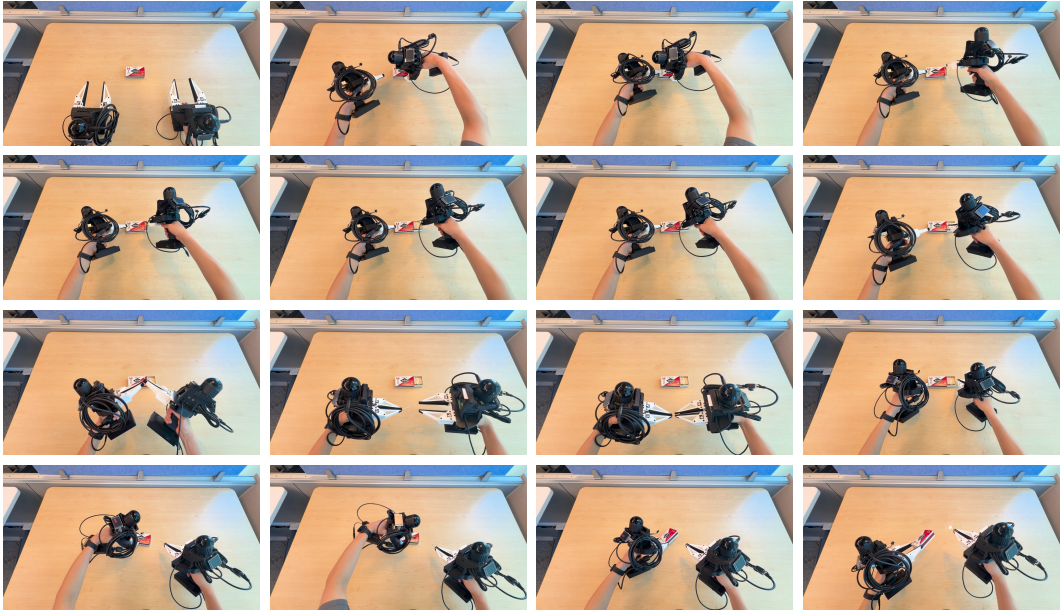}}
  \caption{Bimanual match lighting with \systemname{}. A human operator coordinates the two grippers to open the matchbox, retrieve a match, and strike it. The final frame shows the lit match. Frames are shown in chronological order from left to right and top to bottom.}
  \label{fig:match-lighting}
\end{figure}

Figure~\ref{fig:match-lighting} shows a human operator using both \systemname{} grippers to open a matchbox, retrieve a match, and light it.

\clearpage
\noindent\begin{minipage}{\linewidth}
\subsection{Shirt Folding}
\label{app:shirt-folding}
Figure~\ref{fig:shirt-folding} shows a human operator using both \systemname{} grippers to spread and fold a shirt.

\begingroup
  \centering
  \resizebox{\linewidth}{!}{
\begingroup%
\begin{tikzpicture}[x=1pt,y=-1pt]
\path[use as bounding box] (0,0) rectangle (396,228.000);
\node[anchor=north west,inner sep=0] at (0.000,0.000)
  {\includegraphics[width=96.000000pt]{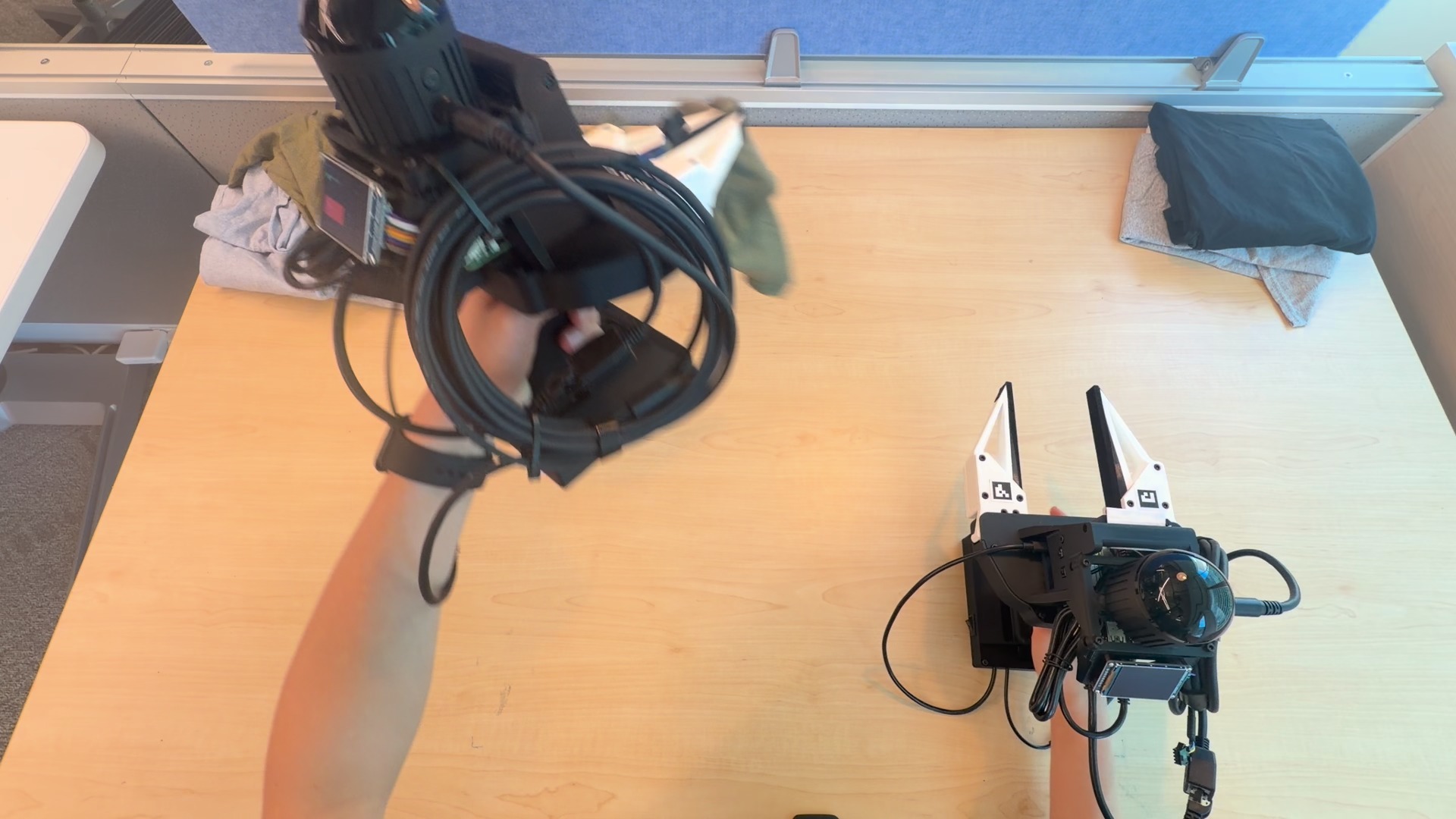}};
\node[anchor=north west,inner sep=0] at (100.000,0.000)
  {\includegraphics[width=96.000000pt]{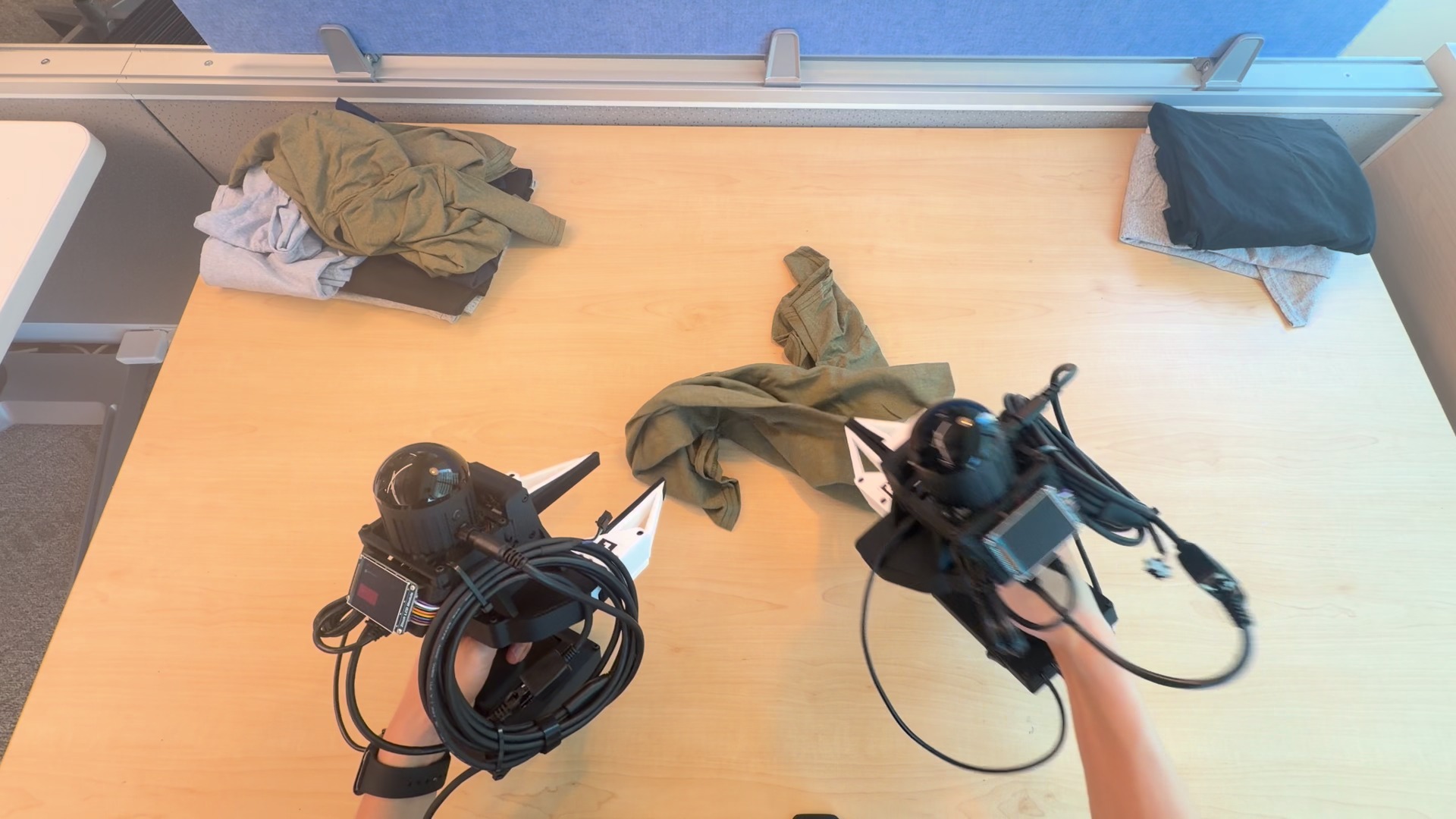}};
\node[anchor=north west,inner sep=0] at (200.000,0.000)
  {\includegraphics[width=96.000000pt]{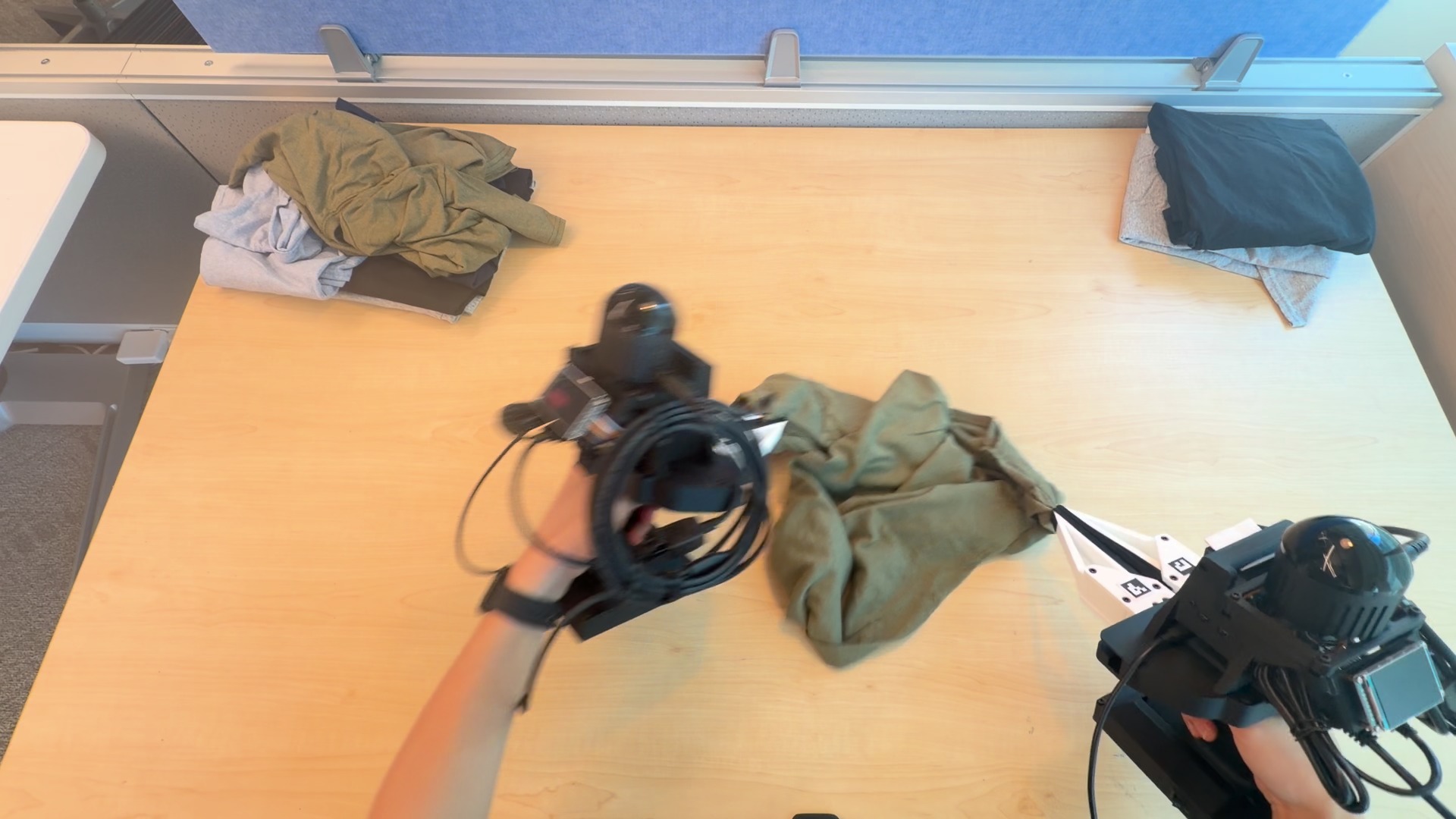}};
\node[anchor=north west,inner sep=0] at (300.000,0.000)
  {\includegraphics[width=96.000000pt]{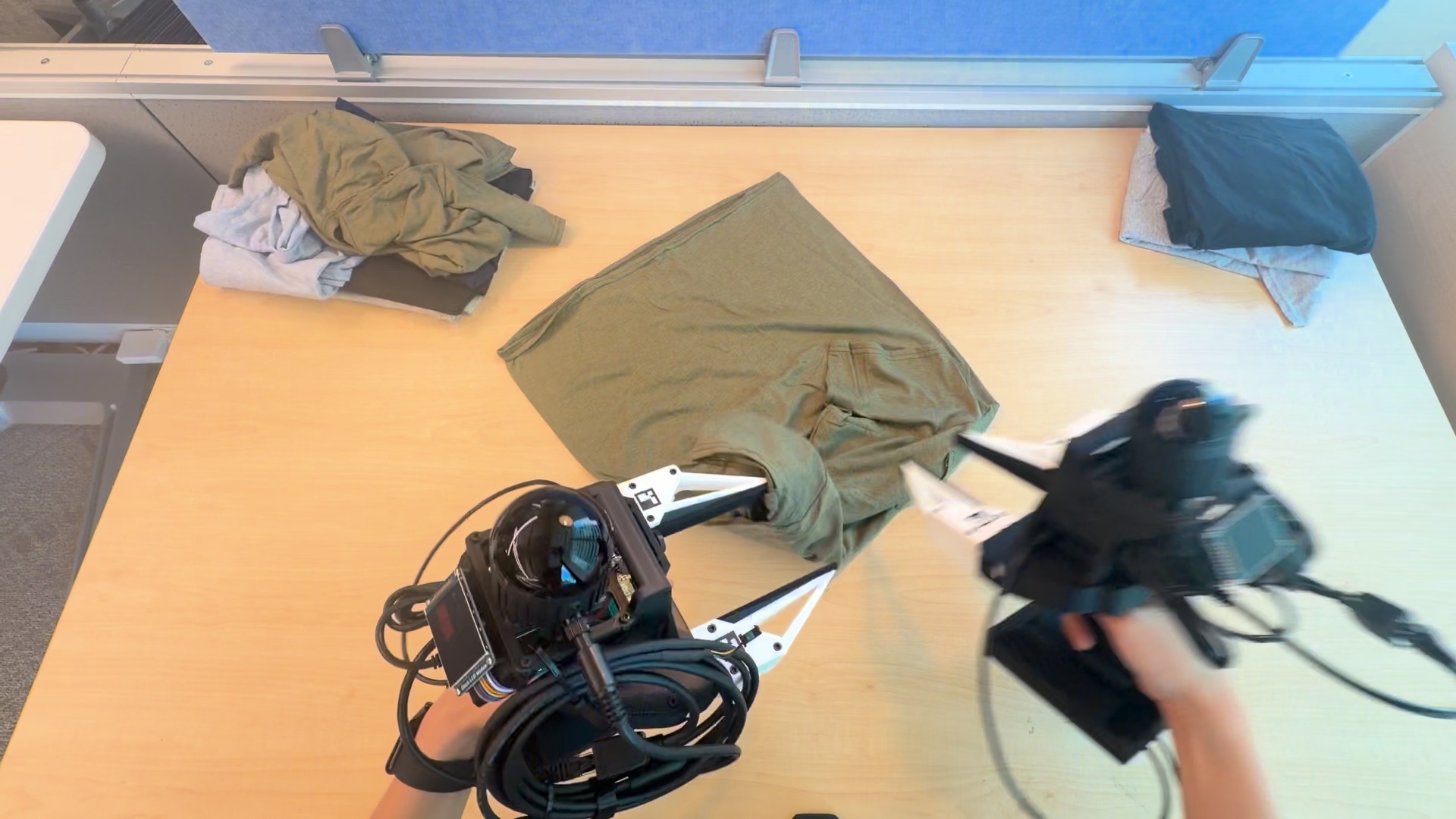}};
\node[anchor=north west,inner sep=0] at (0.000,58.000)
  {\includegraphics[width=96.000000pt]{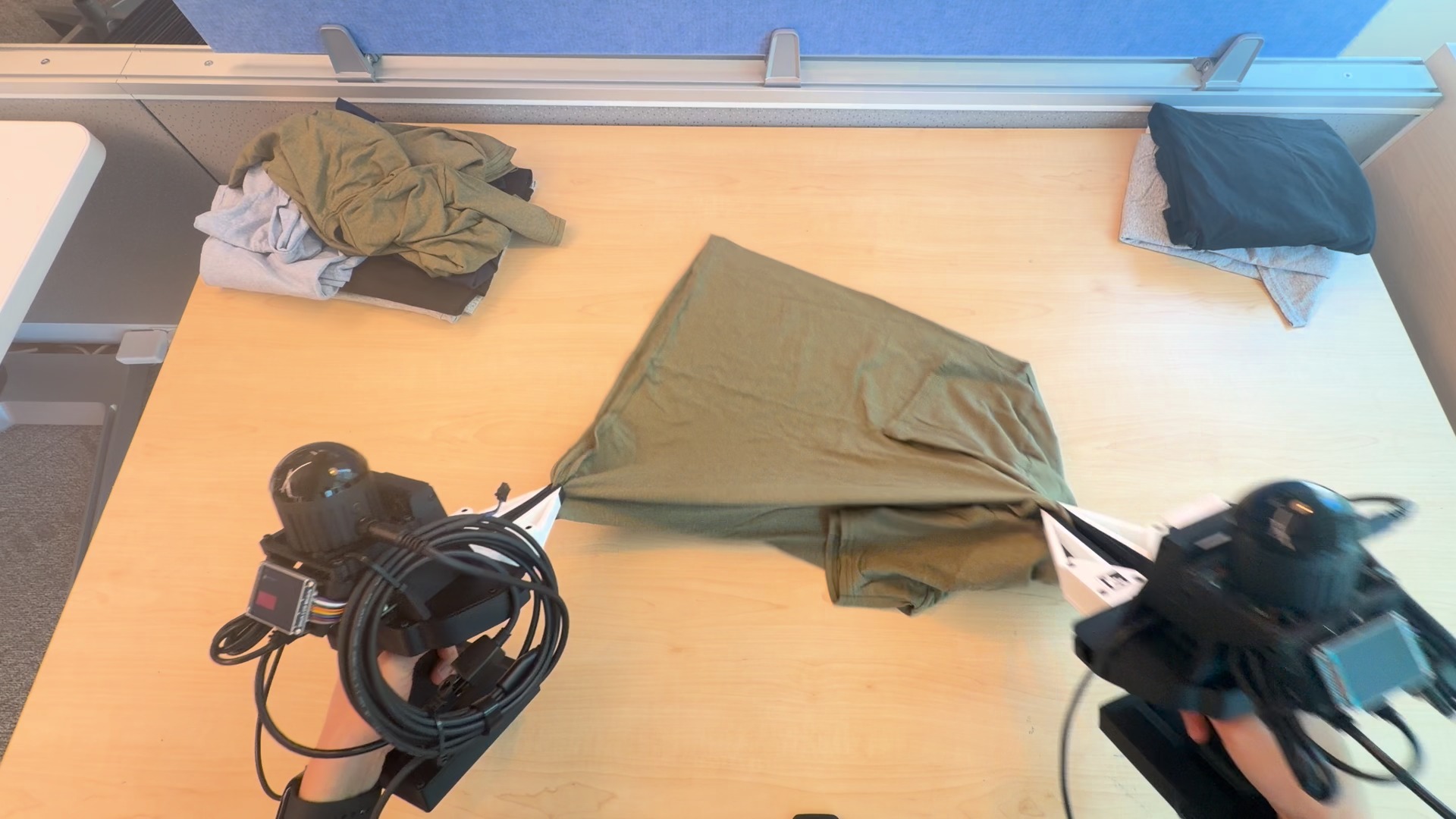}};
\node[anchor=north west,inner sep=0] at (100.000,58.000)
  {\includegraphics[width=96.000000pt]{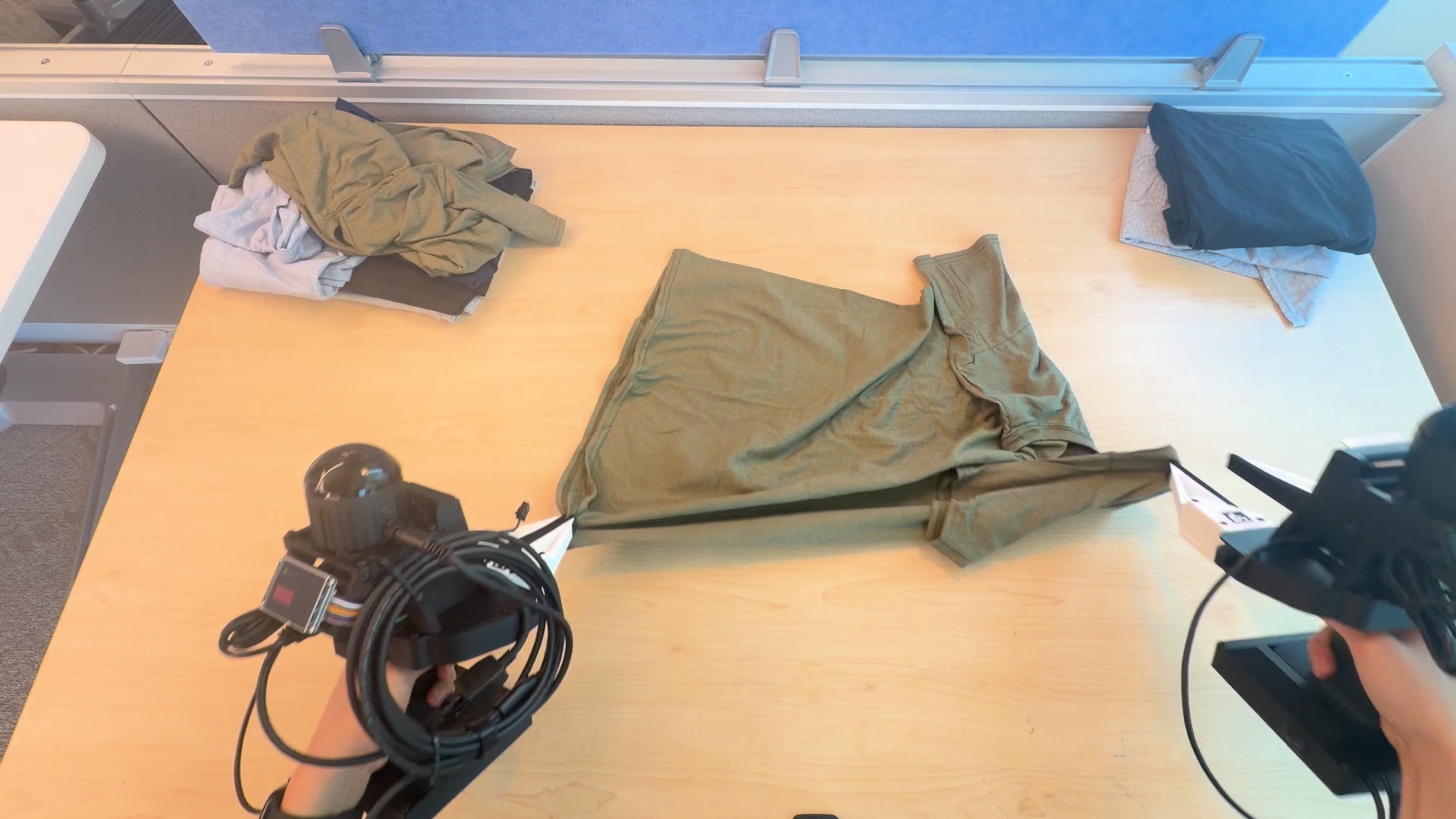}};
\node[anchor=north west,inner sep=0] at (200.000,58.000)
  {\includegraphics[width=96.000000pt]{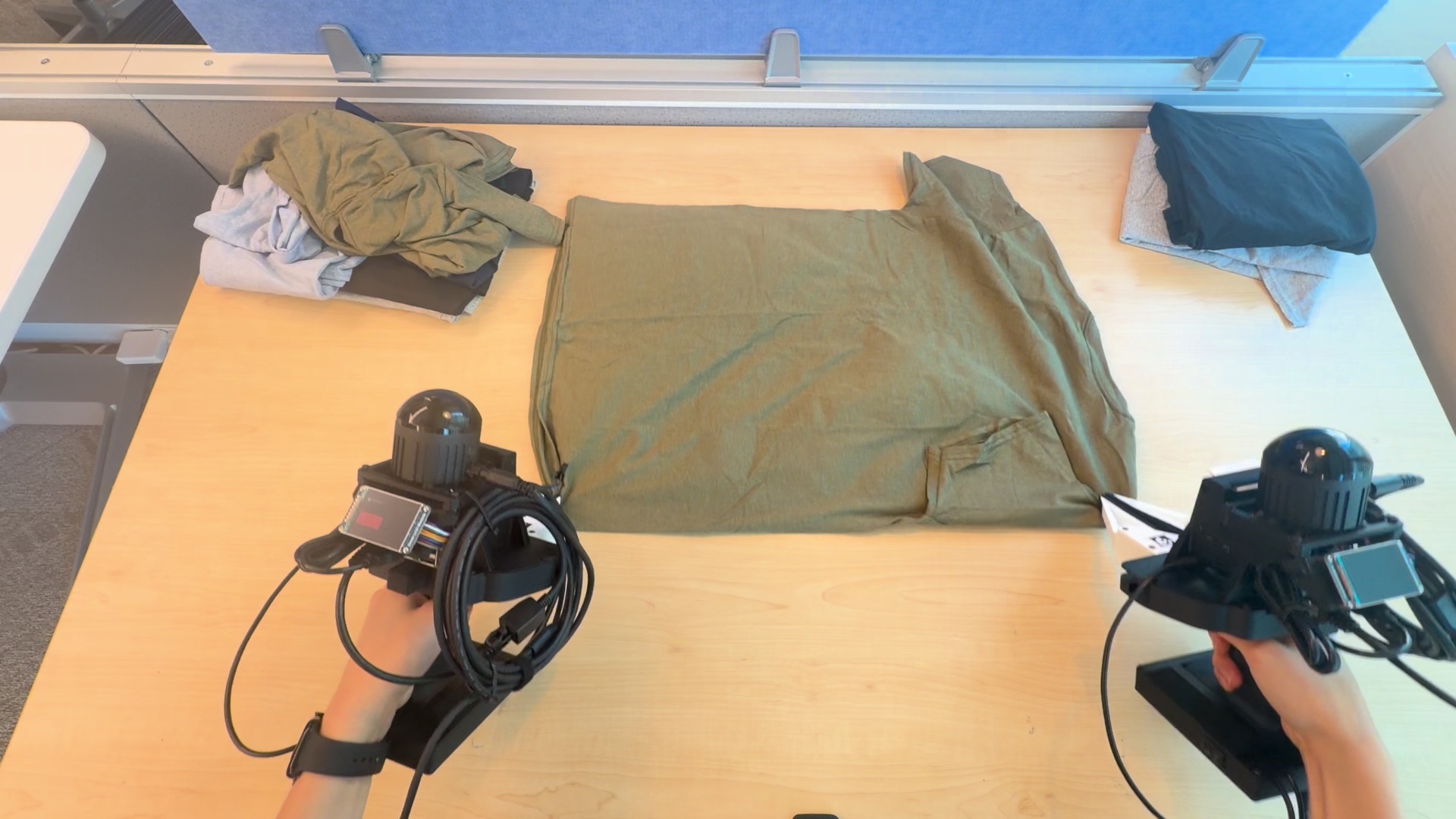}};
\node[anchor=north west,inner sep=0] at (300.000,58.000)
  {\includegraphics[width=96.000000pt]{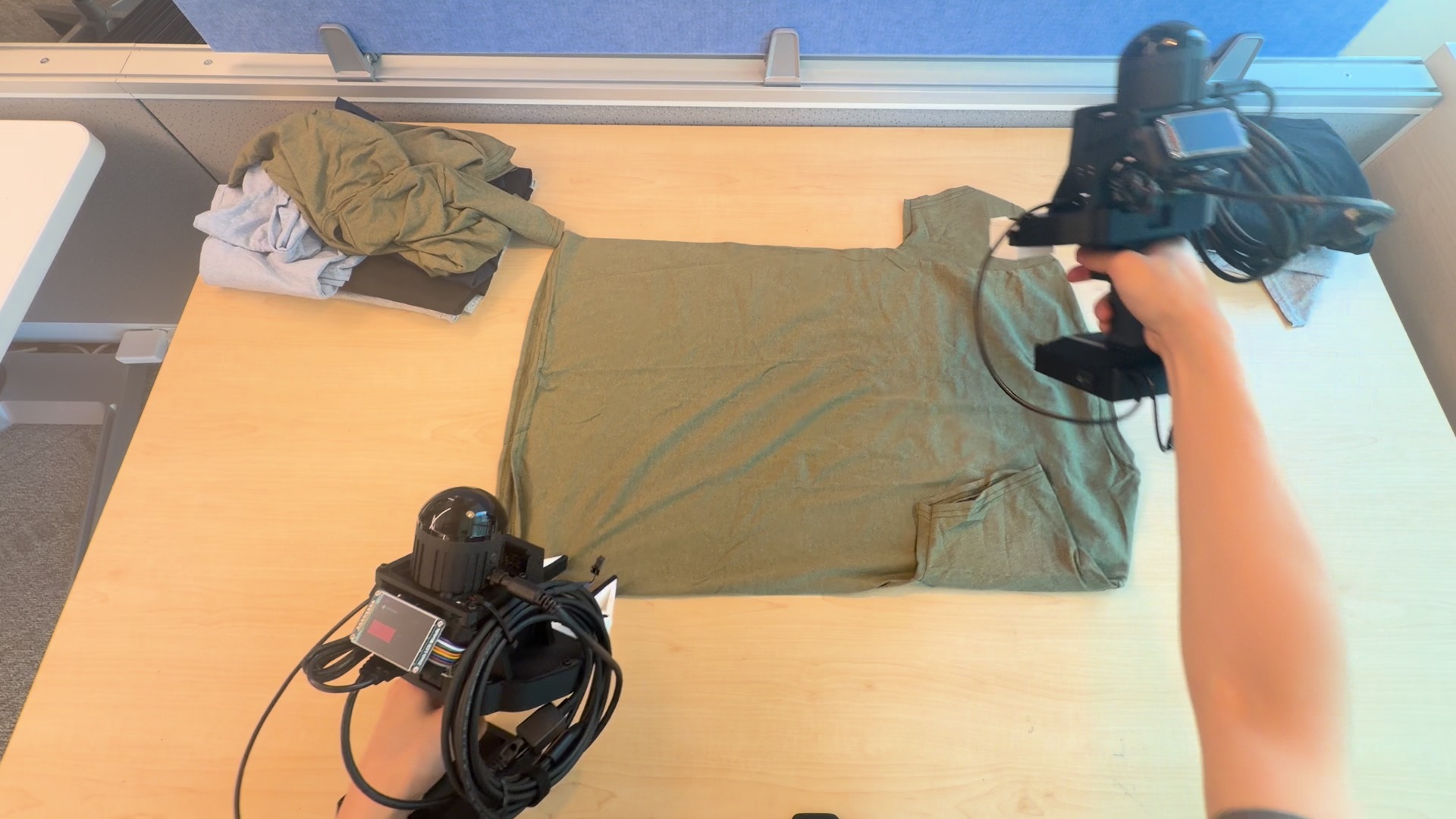}};
\node[anchor=north west,inner sep=0] at (0.000,116.000)
  {\includegraphics[width=96.000000pt]{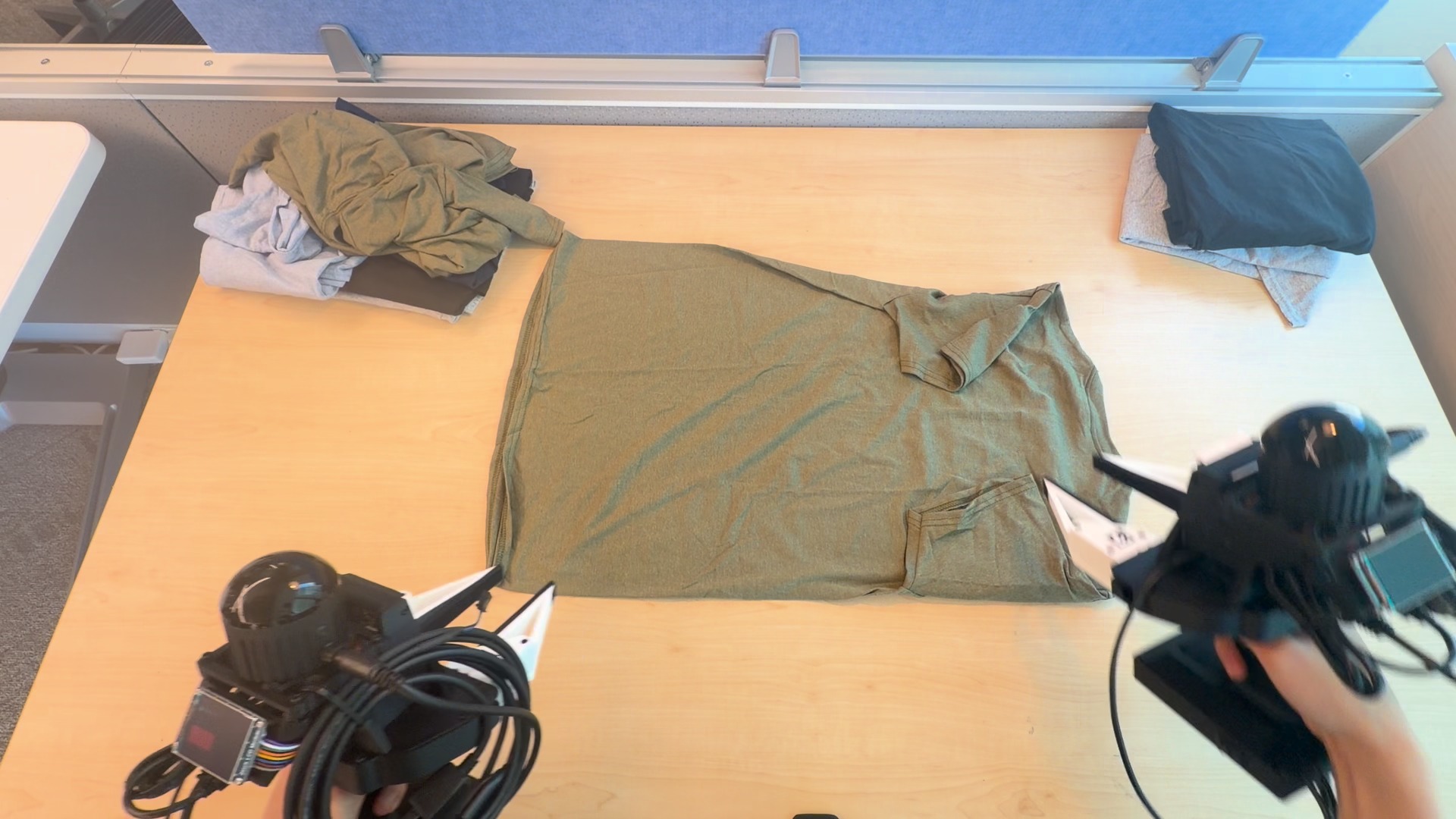}};
\node[anchor=north west,inner sep=0] at (100.000,116.000)
  {\includegraphics[width=96.000000pt]{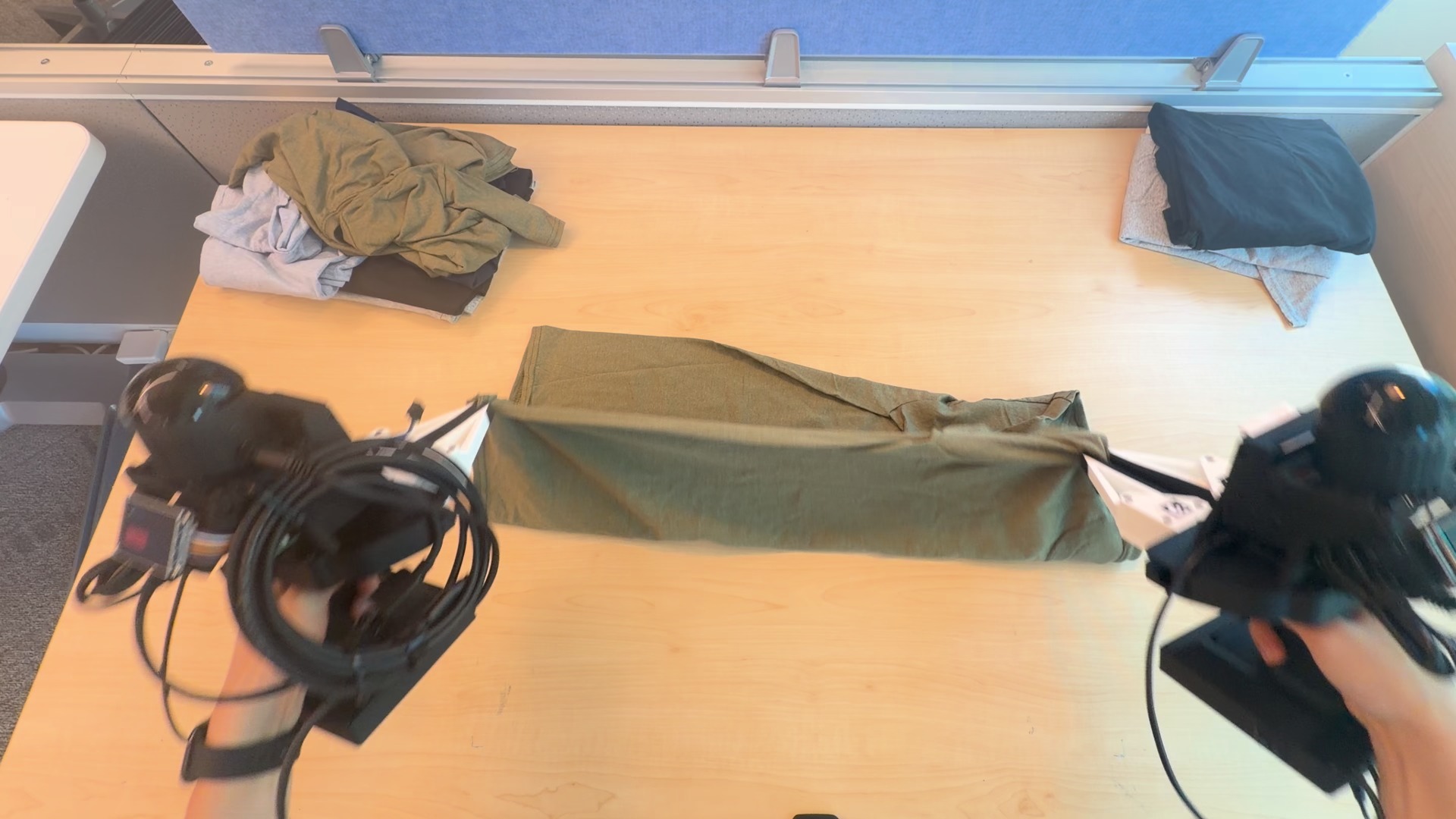}};
\node[anchor=north west,inner sep=0] at (200.000,116.000)
  {\includegraphics[width=96.000000pt]{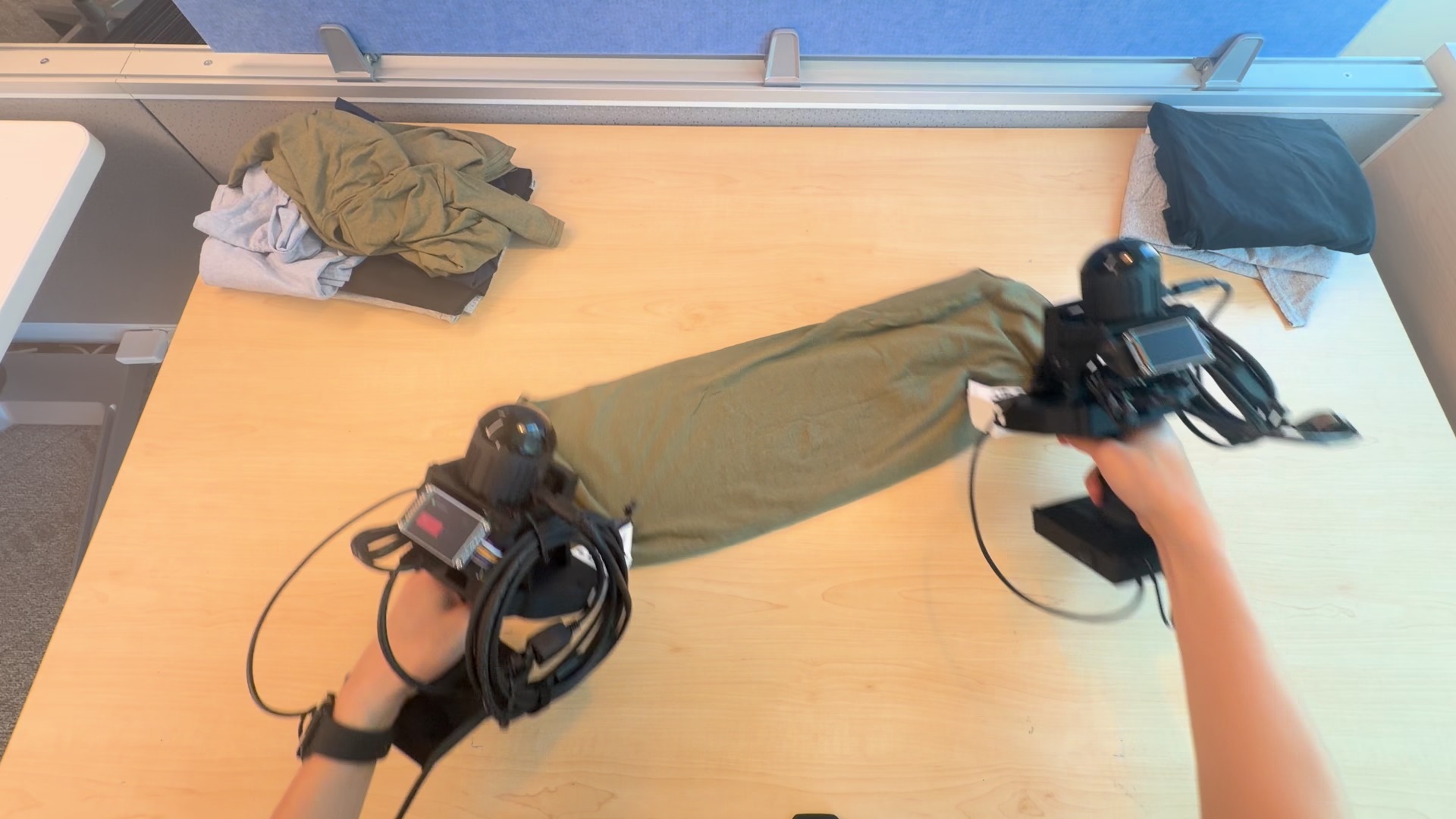}};
\node[anchor=north west,inner sep=0] at (300.000,116.000)
  {\includegraphics[width=96.000000pt]{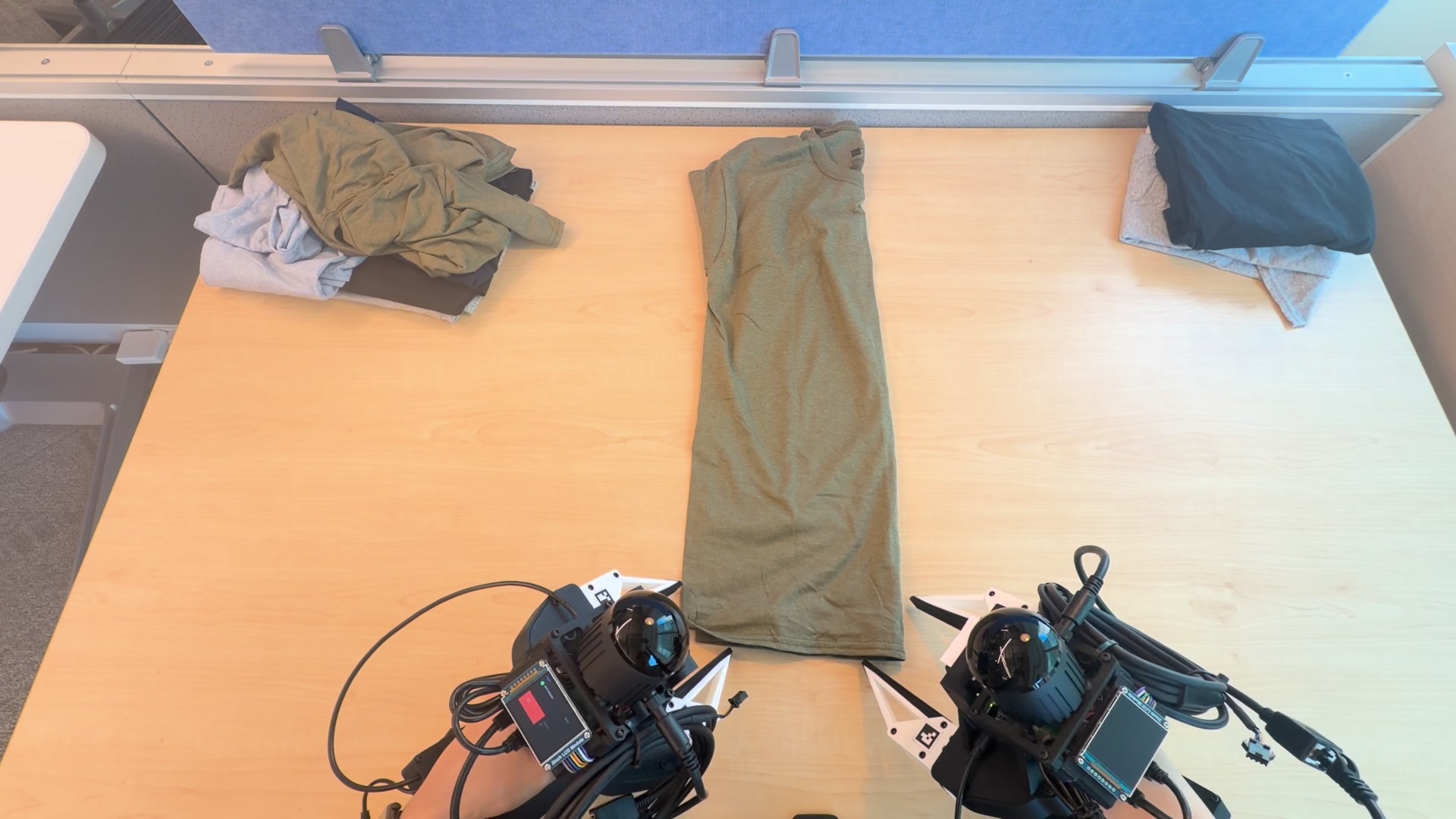}};
\node[anchor=north west,inner sep=0] at (0.000,174.000)
  {\includegraphics[width=96.000000pt]{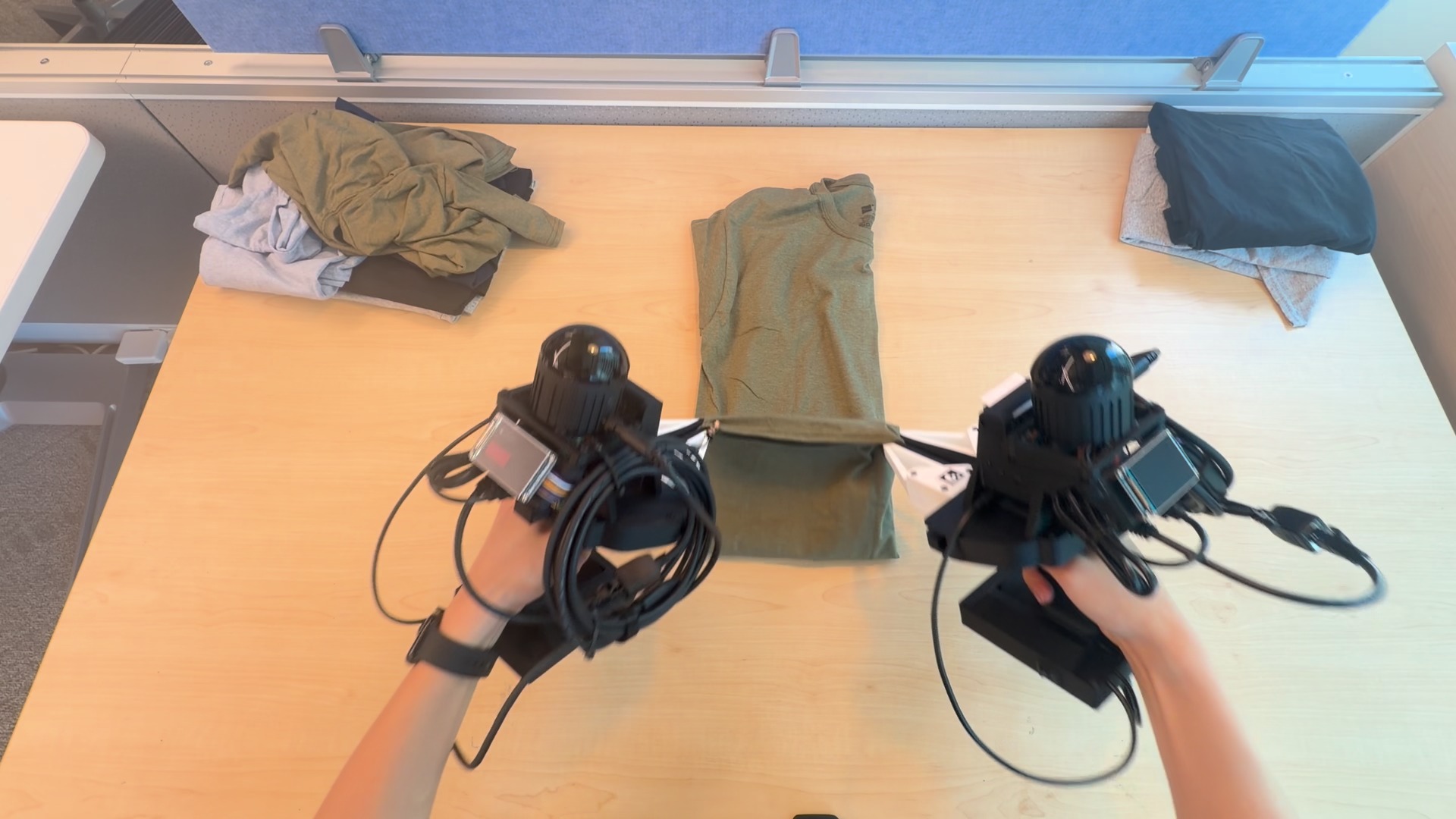}};
\node[anchor=north west,inner sep=0] at (100.000,174.000)
  {\includegraphics[width=96.000000pt]{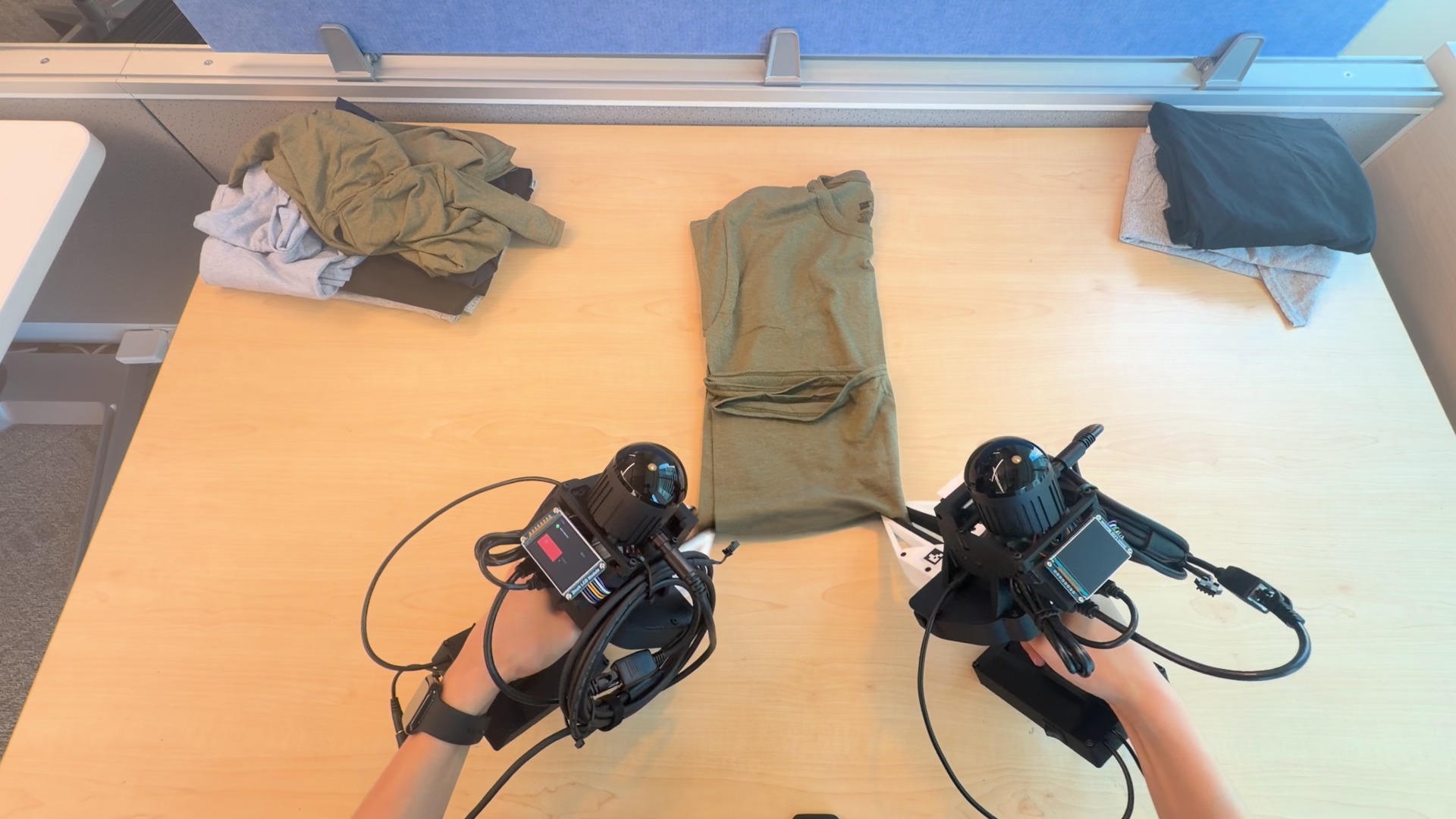}};
\node[anchor=north west,inner sep=0] at (200.000,174.000)
  {\includegraphics[width=96.000000pt]{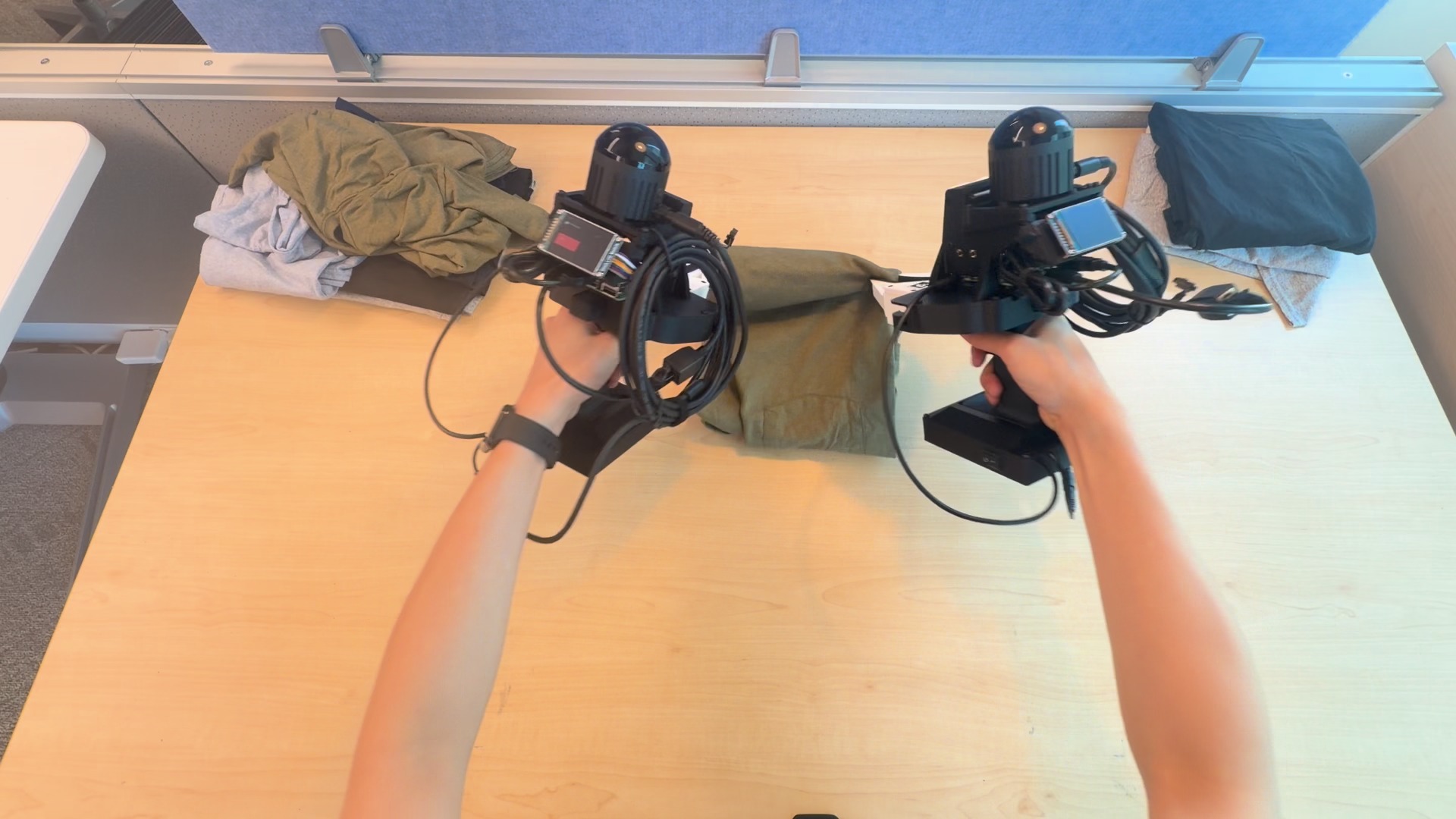}};
\node[anchor=north west,inner sep=0] at (300.000,174.000)
  {\includegraphics[width=96.000000pt]{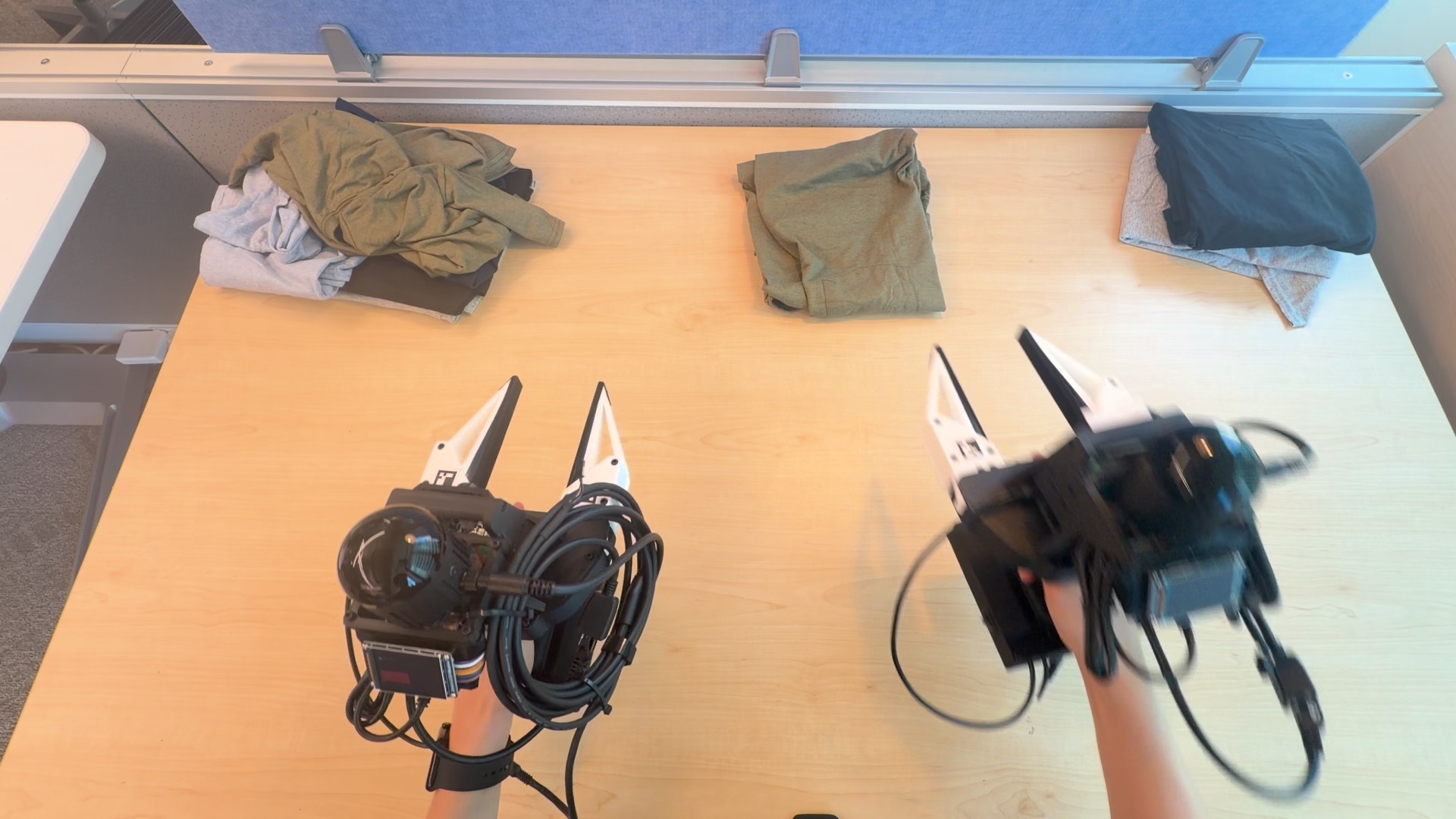}};
\end{tikzpicture}%
\endgroup%
}
  \captionof{figure}{Bimanual shirt folding with \systemname{}. A human operator coordinates both grippers to spread and align the shirt, fold it lengthwise, and complete the final folds. Frames are shown in chronological order from left to right and top to bottom.}
  \label{fig:shirt-folding}
\par\endgroup
\end{minipage}

\FloatBarrier
\clearpage
\section{Autonomous Rollout}
\label{app:autonomous-rollout}
\begin{figure}[!htbp]
  \centering
  \resizebox{\linewidth}{!}{\input{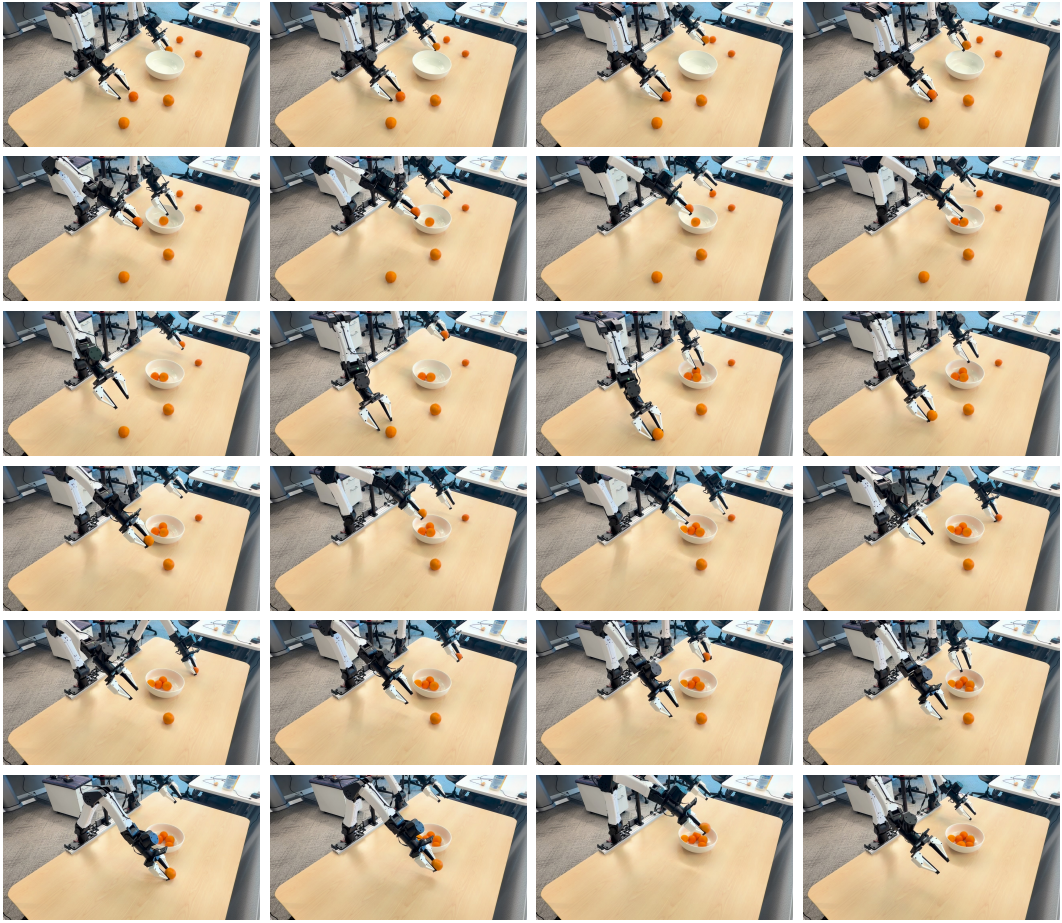}}
  \caption{One autonomous $\pi_{0.5}$ rollout on dual i2RT YAM arms, transferring six oranges from the table into a bowl. Frames are shown in chronological order from left to right and top to bottom.}
  \label{fig:autonomous-rollout}
\end{figure}

Figure~\ref{fig:autonomous-rollout} shows a complete autonomous $\pi_{0.5}$ rollout on the orange collection task.

\FloatBarrier
\section{Teleoperation Setup}
\label{app:teleop-setup}
\begin{figure}[!htbp]
  \centering
  \resizebox{\linewidth}{!}{\input{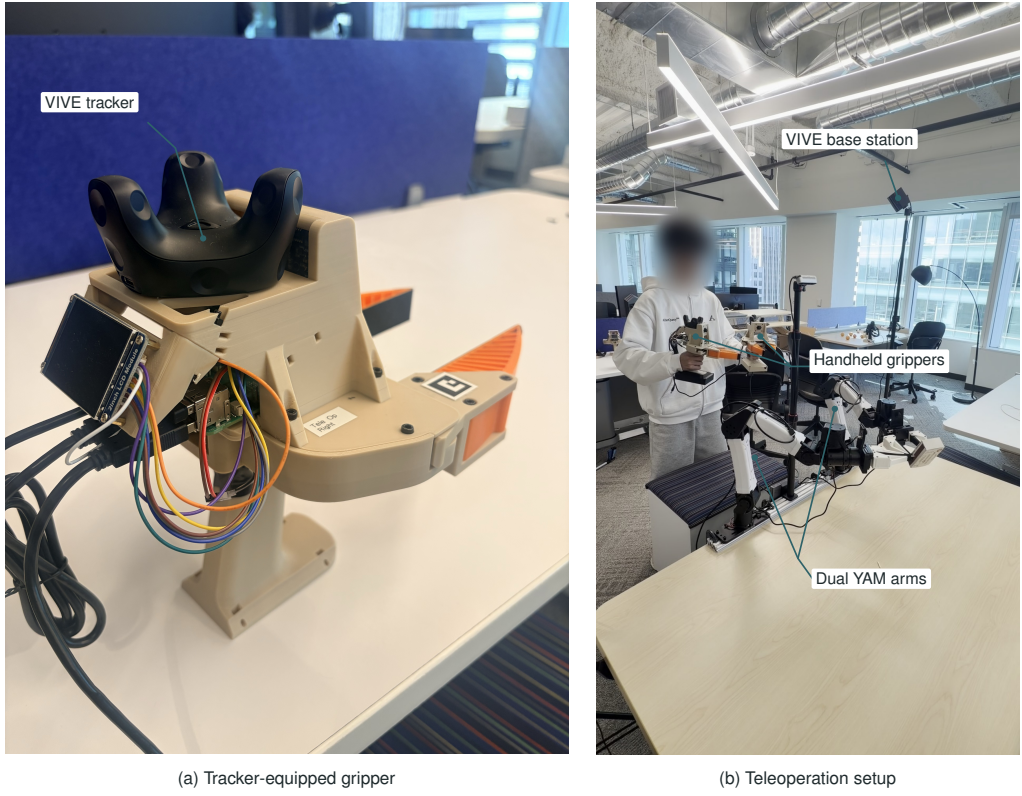}}
  \caption{Teleoperation setup. \textbf{(a)} A VIVE tracker replaces the LiDAR on each handheld gripper. \textbf{(b)} External base-station tracking supports teleoperation of the dual YAM arms.}
  \label{fig:teleop-setup}
\end{figure}

For teleoperation, we replace the LiDAR on each UMI-Live gripper with a VIVE tracker\footnote{\url{https://www.vive.com/us/accessory/tracker3/}} and use Lighthouse base stations\footnote{\url{https://www.vive.com/us/accessory/base-station2/}} for external tracking (Figure~\ref{fig:teleop-setup}). We align the tracking frame with the robot base axes and capture reference poses when control is engaged. Subsequent changes in tracker position and orientation are mapped to Cartesian targets for the corresponding YAM arm and converted into joint commands through IK. Jaw opening is measured from ArUco tags in the wrist-camera images and streamed from the Raspberry Pi to control the corresponding robot gripper, with opening changes scaled by the calibrated gripper span. We release the teleoperation software, setup and calibration instructions in \href{https://github.com/AfterQuery-Research/umi-teleop}{UMI-Teleop}.

\FloatBarrier
\section{Dataset Representation}
\label{app:dataset-representation}

Each UMI-Live demonstration is recorded as an episode with a task instruction $\ell$. For each gripper $i\in\{L,R\}$, we define its initial TCP pose as zero and express all subsequent TCP positions and orientations relative to this fixed reference. At frame $t$, we store a wrist image $I_t^i$ and the state
\begin{equation}
\mathbf{s}_t^i=
[x_t^i,y_t^i,z_t^i,\;
\phi_{x,t}^i,\phi_{y,t}^i,\phi_{z,t}^i,\;
g_t^i]\in\mathbb R^7,
\end{equation}
where $\mathbf{p}_t^i = (x_t^i,y_t^i,z_t^i)^\top\in\mathbb R^3$ is position in meters, $\bm{\phi}_t^i = (\phi_{x,t}^i,\phi_{y,t}^i,\phi_{z,t}^i)^\top\in\mathbb R^3$ is a rotation vector in radians, and $g_t^i\in[0,1]$ is the normalized gripper opening, such that
\begin{equation}
g_t^i=\operatorname{clip}\!\left(
\frac{w_t^i-w_{\mathrm{closed}}^i}{w_{\mathrm{open}}^i-w_{\mathrm{closed}}^i},\,0,\,1
\right),
\label{eq:dataset-gripper-normalization}
\end{equation}
where $w_t^i$ are the physical jaw widths and $w_\text{open}, w_\text{closed}$ are the calibrated open and closed endpoints.

The bimanual state concatenates both hands,
\begin{equation}
\mathbf{s}_t^{\mathrm{bi}}=[\mathbf{s}_t^L;\,\mathbf{s}_t^R]\in\mathbb R^{14}.
\label{eq:dataset-state}
\end{equation}
An episode with $N$ frames is represented as the demonstration trajectory:
\begin{equation}
\mathcal T=
\left[
\left(I_t^L,I_t^R,\mathbf{s}_t^{\mathrm{bi}},\ell\right)
\right]_{t=0}^{N-1}.
\end{equation}
We denote \(\mathcal D=\{\mathcal T^{(e)}\}_{e=1}^{M}\) as a dataset of \(M\) demonstration trajectories, and $\mathcal T=(\mathcal T_0,\ldots,\mathcal T_{N-1})$.

\begin{figure}[!htbp]
\centering
\includegraphics[width=\linewidth]{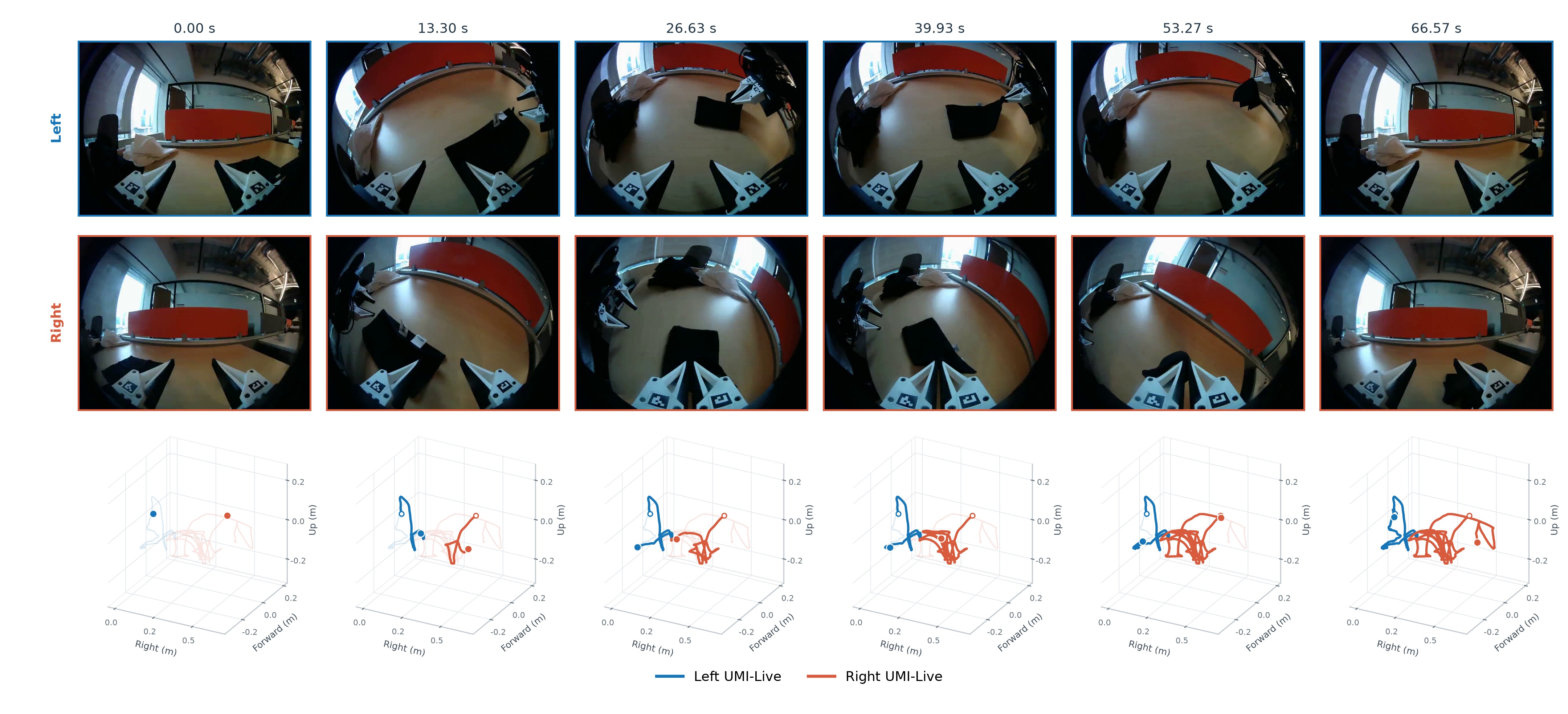}
\caption{Wrist images and reconstructed TCP position trajectories during bimanual folding. Each column shows the left wrist view (top), right wrist view (middle), and the two TCP position traces in a common plotting frame (bottom).}
\label{fig:dataset-representation}
\end{figure}

Figure~\ref{fig:dataset-representation} shows the left and right wrist views and corresponding TCP position trajectories during an episode.

\FloatBarrier
\section{Policy Training}
\label{app:policy-training}

We fine-tune MolmoAct2~\citep{fang2026molmoact2} and $\pi_{0.5}$~\citep{intelligence2025pi_} to predict bimanual TCP trajectories. We first define training on UMI demonstrations, then extend the same representation and losses to mixtures with robot teleoperation data.

\subsection{Training on UMI Demonstrations}
\label{app:training-representation}
We uniformly shuffle training frames across \(\mathcal D\) and retrieve the corresponding trajectory \(\mathcal T\) for each frame \(t\). The current entry \(\mathcal T_t\) provides the policy observation, while the next \(H\) entries provide the future states used to construct supervision,
\[
o_t=(I_t^L,I_t^R,g_t^L,g_t^R,\ell),
\qquad
Y_t=(\mathbf s_{t+1}^{\mathrm{bi}},\ldots,\mathbf s_{t+H}^{\mathrm{bi}}),
\]
where \(H\) is the action horizon. We omit TCP pose and joint information from the policy inputs. Both \(\pi_{0.5}\) and MolmoAct2 inherit visual representations from large-scale vision-language pretraining. Together with the broad workspace coverage of our ultra-wide wrist cameras, these representations provide visual context for learning bimanual coordination.

For each gripper \(i\in\{L,R\}\), the stored state is
\[
\mathbf s_t^i=[\mathbf p_t^i;\boldsymbol\phi_t^i;g_t^i]\in\mathbb R^7.
\]
We reconstruct the TCP pose as
\[
R_t^i=\exp([\boldsymbol\phi_t^i]_\times) \in \mathbb{R}^{3\times3},
\qquad
T_t^i=
\begin{bmatrix}
R_t^i&\mathbf p_t^i\\
\mathbf 0^\top&1
\end{bmatrix}
\in SE(3),
\]
where \([\cdot]_\times\) denotes the skew-symmetric matrix associated with a vector.

We express each future pose relative to the current pose,
\[
\Delta T_{t,k}^i=(T_t^i)^{-1}T_{t+k}^i,
\qquad k=1,\ldots,H.
\]
Its translation and rotation are
\[
\Delta\mathbf p_{t,k}^i
=(R_t^i)^\top(\mathbf p_{t+k}^i-\mathbf p_t^i),
\qquad
\Delta R_{t,k}^i=(R_t^i)^\top R_{t+k}^i.
\]
We encode rotation using the first two rows of the rotation matrix to construct the 6D rotation vector~\citep{zhou2019continuity},
\[
\mathbf r_6(R)
=[R_{11},R_{12},R_{13},R_{21},R_{22},R_{23}]^\top.
\]
The resulting action target for each gripper is
\[
\mathbf a_{t,k}^i
=
[\Delta\mathbf p_{t,k}^i;
\mathbf r_6(\Delta R_{t,k}^i);
g_{t+k}^i]
\in\mathbb R^{10}.
\]
Concatenating the two grippers produces the bimanual action chunk
\[
A_t=
\begin{bmatrix}
[\mathbf a_{t,1}^L;\mathbf a_{t,1}^R]^\top\\
\vdots\\
[\mathbf a_{t,H}^L;\mathbf a_{t,H}^R]^\top
\end{bmatrix}
\in\mathbb R^{H\times20}.
\]
All future poses share the current frame \(t\) as their reference, while gripper openings remain absolute targets.

We train the conditional action model using flow matching~\citep{lipman2023flowmatching}. Given Gaussian noise \(\epsilon\sim\mathcal N(0,I)\) and flow time \(\tau\), we construct
\[
X_{t,\tau}=(1-\tau)A_t+\tau\epsilon.
\]
The model \(f_\theta(o_t,X_{t,\tau},\tau)\) predicts a velocity field with the same dimensions as the action chunk. We minimize
\[
\mathcal L_{\mathrm{flow}}
=
\mathbb E_{\mathcal T,t,\epsilon,\tau}
\left[
\frac{1}{20H}
\left\|
f_\theta(o_t,X_{t,\tau},\tau)
-(\epsilon-A_t)
\right\|_F^2
\right].
\]

\FloatBarrier
\subsection{Training with Teleoperation Data}
\label{app:mixed-data-training}

We supplement UMI demonstrations with robot teleoperation data to expose the policies to observations and motions collected on the target robot. Each teleoperation episode contains the two wrist images, measured arm joint angles \(\mathbf q_t^i\), gripper openings, and a task instruction \(\ell\). For each arm \(i\in\{L,R\}\), we recover the TCP pose through forward kinematics,
\[
T_t^{i,\mathrm{robot}}=\operatorname{FK}_i(\mathbf q_t^i).
\]
We normalize gripper openings using the calibrated closed and open endpoints and construct future pose targets relative to the current robot pose,
\[
\Delta T_{t,k}^i
=
(T_t^{i,\mathrm{robot}})^{-1}T_{t+k}^{i,\mathrm{robot}},
\qquad k=1,\ldots,H.
\]
The translation, six-dimensional rotation encoding, and future gripper opening are assembled into \(A_t\) using the same construction as for UMI demonstrations.
Both sources therefore provide the same observation format \(o_t\) and action representation \(A_t\). Joint angles are used to reconstruct robot trajectories and are omitted from the policy inputs. Let \(\mathcal D_{\mathrm{teleop}}\) denote the teleoperation trajectories. We fine-tune on the combined dataset
\[
\mathcal D_{\mathrm{mix}}
=
\mathcal D\cup\mathcal D_{\mathrm{teleop}},
\]
using the training objective defined above. The UMI and robot teleoperation shares sum to 100\% of the fine-tuning mixture. Section~\ref{sec:experiments} reports the percentage splits used for each policy.

\FloatBarrier
\subsection{Training Setup}
\label{app:training-setup}

We initialize $\pi_{0.5}$ from the released base model and MolmoAct2 from \verb|MolmoAct2-BimanualYAM|. Both models are fine-tuned for 12,000 optimizer updates using AdamW~\citep{loshchilov2019adamw}, 600 warmup updates, and cosine learning-rate decay to 0.1 of each parameter group's peak rate. We use an action horizon of $H=24$ at 30\,Hz, gradient clipping at global norm 1. We update all $\pi_{0.5}$ parameters. For MolmoAct2, we train LoRA adapters in the vision-language backbone and the full action expert, and we keep the remaining backbone weights and input embeddings fixed. MolmoAct2 combines flow matching with discrete action-token supervision. Tables~\ref{tab:training-pi05} and~\ref{tab:training-molmoact2} summarize the settings for these fresh-base runs.


\begin{table}[!htbp]
\centering
\begin{tabular*}{\linewidth}{@{\extracolsep{\fill}}ll@{}}
\toprule
Hyperparameter & Value \\
\midrule
Global batch size & 32 \\
Parameters updated & Full model \\
Peak learning rate & $2.5\times10^{-5}$ \\
AdamW $\beta$ & $(0.9,0.95)$ \\
AdamW $\epsilon$ & $10^{-8}$ \\
Weight decay & 0.01 \\
Model precision & FP32 \\
Flow samples per training query & 1 \\
\bottomrule
\end{tabular*}
\caption{$\pi_{0.5}$ fine-tuning hyperparameters for the fresh-base runs.}
\label{tab:training-pi05}
\end{table}

\begin{table}[!htbp]
\centering
\begin{tabular*}{\linewidth}{@{\extracolsep{\fill}}ll@{}}
\toprule
Hyperparameter & Value \\
\midrule
Global batch size & 32 \\
Parameters updated & VLM LoRA adapters and full action expert \\
Peak adapter learning rate & $5\times10^{-5}$ \\
Peak action expert learning rate & $10^{-4}$ \\
AdamW $\beta$ & $(0.9,0.95)$ \\
AdamW $\epsilon$ & $10^{-6}$ \\
Weight decay & 0 \\
LoRA rank / $\alpha$ / dropout & 64 / 16 / 0.05 \\
Language-model residual dropout & 0.1 \\
Model precision & BF16 \\
Flow samples per training query & 8 \\
\bottomrule
\end{tabular*}
\caption{MolmoAct2 fine-tuning hyperparameters for the fresh-base run.}
\label{tab:training-molmoact2}
\end{table}

\FloatBarrier
\section{Policy Inference}
\label{app:policy-inference}

At each policy query \(t\), the robot provides the observation
\[
o_t=(I_t^L,I_t^R,g_t^L,g_t^R,\ell).
\]
Conditioned on \(o_t\), we generate a bimanual action chunk by integrating the learned flow field. We express both policies using the flow-time convention defined during training, with Gaussian noise at \(\tau=1\) and actions at \(\tau=0\).
We initialize \(X_{t,1}=\epsilon\), where \(\epsilon\sim\mathcal N(0,I)\) has shape \(H\times20\), and solve
\[
\frac{dX_{t,\tau}}{d\tau}
=
f_\theta(o_t,X_{t,\tau},\tau)
\]
from \(\tau=1\) to \(\tau=0\). Here, \(\tau\) denotes flow time, distinct from the robot timestep \(t\). The observation remains fixed while the entire action chunk is refined.
Using \(K\) Euler steps with \(\tau_j=1-j/K\), we compute
\[
X_{t,\tau_{j+1}}
=
X_{t,\tau_j}
-
\frac{1}{K}
f_\theta(o_t,X_{t,\tau_j},\tau_j),
\qquad j=0,\ldots,K-1.
\]
The final estimate gives the predicted action chunk
\[
\widehat A_t=X_{t,0}\in\mathbb R^{H\times20}.
\]
Each predicted action contains a relative translation, a six-dimensional rotation representation, and an absolute gripper opening for each arm. We recover the relative rotation through Gram-Schmidt orthonormalization and combine it with the predicted translation to form \(\widehat{\Delta T}_{t,k}^i\).
At the query timestep, the robot computer records the measured joint angles \(\mathbf q_t^i\) and computes the TCP pose
\[
T_t^{i,\mathrm{robot}}
=
\operatorname{FK}_i(\mathbf q_t^i).
\]
These joint measurements are used by the execution controller and are omitted from the policy inputs. Each relative prediction is mapped to a robot target pose,
\[
\widehat T_{t,k}^{i,\mathrm{target}}
=
T_t^{i,\mathrm{robot}}\widehat{\Delta T}_{t,k}^i,
\qquad k=1,\ldots,H.
\]
Inverse kinematics converts this target into joint commands,
\[
\widehat{\mathbf q}_{t,k}^i
=
\operatorname{IK}_i
\left(
\widehat T_{t,k}^{i,\mathrm{target}};
\mathbf q_t^i
\right),
\]
using the measured query joints as the initial configuration. Predicted gripper openings are mapped to calibrated robot gripper commands.
All targets within a chunk share the pose measured at its query. The robot executes actions while the server processes subsequent observations. Returned predictions are aligned with the robot’s execution timeline, and elapsed predictions are discarded. Each new query supplies updated observations and a fresh pose reference.

\subsection{Dual-YAM Hardware Integration}
\label{app:dual-yam-integration}

We extend LeRobot's existing client--server inference architecture to support our dual-YAM setup. We run policy inference on a remote server and execute the returned action chunks through a local robot client. We run each arm in a separate worker process that reads joint and gripper feedback and dispatches commands through I2RT. We assign left and right commands a common scheduled execution time to coordinate bimanual motion. We use I2RT for forward and inverse kinematics, and we verify each inverse-kinematics solution against pose-error tolerances, joint limits, and a bound on joint displacement. We map predicted gripper openings to hardware commands using saved gripper calibrations. We also provide an action-probe mode that logs predicted commands without dispatching policy actions.

\subsection{Kinematic Model}
\label{app:yam-kinematics}

We retain the six-joint kinematic structure and arm geometry of the default YAM configuration. We adapt the model to our replacement gripper by updating the TCP position and orientation through a fixed flange-to-TCP transform.

We denote a transformation from frame \(a\) to frame \(b\) by \(T_a^b\). We describe the arm using the classical Denavit--Hartenberg convention
\[
T_j^{j-1}(q_j)
=
R_z(q_j+\theta_j^{\mathrm{offset}})
\,T_z(d_j)\,T_x(a_j)\,R_x(\alpha_j),
\]
where \(q_j\) is the angle of joint \(j\). We use the parameters in Table~\ref{tab:yam-dh} from the I2RT YAM model,\footnote{\href{https://github.com/AfterQuery-Research/i2rt/blob/7ed46f4e4e316133a0c39aa6cf34a73d2718e850/i2rt/robot_models/arm/yam/README.md}{I2RT YAM model, revision \texttt{7ed46f4}.}} with distances in meters and angles in radians.

\begin{table}[!htbp]
\centering
\begin{tabular}{ccccc}
\toprule
Joint \(j\) & \(\theta_j^{\mathrm{offset}}\) & \(d_j\) & \(a_j\) & \(\alpha_j\) \\
\midrule
1 & \(0\) & \(0.1135\) & \(0.02\) & \(-\pi/2\) \\
2 & \(-\pi\) & \(0\) & \(0.264\) & \(-\pi\) \\
3 & \(-2.901418\) & \(0\) & \(0.252239\) & \(0\) \\
4 & \(-0.240173\) & \(0\) & \(0.073999\) & \(\pi/2\) \\
5 & \(\pi/2\) & \(0\) & \(0\) & \(\pi/2\) \\
6 & \(\pi/2\) & \(0.0356\) & \(0\) & \(-\pi\) \\
\bottomrule
\end{tabular}
\caption{Classical Denavit--Hartenberg parameters for the YAM arm.}
\label{tab:yam-dh}
\end{table}

We denote the base, flange, and TCP frames by \(B\), \(F\), and \(G\), with joint frames \(0=B\) and \(6=F\). We place the TCP frame origin at the center of the gripper tips, 220\,mm along the negative \(z\)-axis of the flange frame, and orient its axes to match the UMI convention. We use the fixed transform
\[
T_G^F=
\begin{bmatrix}
-1&0&0&0\\
0&1&0&0\\
0&0&-1&-0.220\\
0&0&0&1
\end{bmatrix}.
\]
For either arm, we obtain the TCP pose as
\[
T_G^B(\mathbf q)
=
\left(\prod_{j=1}^{6}T_j^{j-1}(q_j)\right)T_G^F.
\]
We incorporate this TCP transform directly into the I2RT robot model and use its MJCF-based implementation for forward and inverse kinematics during execution.

\end{document}